\documentclass[sigconf]{acmart}
\usepackage{multirow}
\usepackage[table,xcdraw]{xcolor}
\usepackage{bm}
\usepackage{makecell}
\usepackage{tcolorbox}

\AtBeginDocument{%
  }
    
\renewcommand\footnotetextcopyrightpermission[1]{} 
\setcopyright{none}

\begin{document}

\title[Lost in the Hype]{Lost in the Hype: Revealing and Dissecting the Performance Degradation of Medical Multimodal Large Language Models in Image Classification}

\author{Xun Zhu}
\affiliation{%
  \institution{Department of Electronic Engineering\\ Tsinghua University}
  \city{Beijing}
  \country{China}
  }
\email{zhu-x24@mails.tsinghua.edu.cn}

\author{Fanbin Mo}
\affiliation{%
  \institution{School of Artificial Intelligence \\ BUPT}
  \city{Beijing}
  \country{China}
  }
\email{mofanbin@bupt.edu.cn}

\author{Xi Chen}
\affiliation{%
  \institution{Department of Electronic Engineering \\ Tsinghua University}
  \city{Beijing}
  \country{China}
  }
\email{chenxi24@mails.tsinghua.edu.cn}

\author{Kaili Zheng}
\affiliation{%
  \institution{Department of Electronic Engineering \\ Tsinghua University}
  \city{Beijing}
  \country{China}
  }
\email{zkl25@mails.tsinghua.edu.cn}

\author{Shaoshuai Yang}
\affiliation{%
  \institution{Department of Electronic Engineering \\ Tsinghua University}
  \city{Beijing}
  \country{China}
  }
\email{yangss24@mails.tsinghua.edu.cn}

\author{Yiming Shi}
\affiliation{%
  \institution{Department of Electronic Engineering \\ Tsinghua University}
  \city{Beijing}
  \country{China}
  }
\email{sym23@mails.tsinghua.edu.cn}

\author{Jian Gao}
\affiliation{%
  \institution{College of AI \\ Tsinghua University}
  \city{Beijing}
  \country{China}
  }
\email{gaojian21@mails.tsinghua.edu.cn}

\author{Miao Li}
\authornote{Corresponding authors.} 
\affiliation{%
  \institution{Department of Electronic Engineering \\ Tsinghua University}
  \city{Beijing}
  \country{China}
  }
\email{miao-li@tsinghua.edu.cn}

\author{Ji Wu}
\authornotemark[1]
\affiliation{%
  \institution{Department of Electronic Engineering \\ College of AI, Tsinghua University}
  \city{Beijing}
  \country{China}
  }
\email{wuji_ee@tsinghua.edu.cn}

\ccsdesc[500]{Computing methodologies~Machine learning}
\ccsdesc[500]{Computing methodologies~Computer vision}
\ccsdesc[500]{Networks~Network performance evaluation}
\ccsdesc[500]{Applied computing~Life and medical sciences}

\begin{abstract}
The rise of multimodal large language models (MLLMs) has sparked an unprecedented wave of applications in the field of medical imaging analysis.
However, as one of the earliest and most fundamental tasks integrated into this paradigm, medical image classification reveals a sobering reality: state-of-the-art medical MLLMs consistently underperform compared to traditional deep learning models, despite their overwhelming advantages in pre-training data and model parameters.
This paradox prompts a critical rethinking: where exactly does the performance degradation originate?
In this paper, we conduct extensive experiments on 14 open-source medical MLLMs across three representative image classification datasets.
Moving beyond superficial performance benchmarking, we employ feature probing to track the information flow of visual features module-by-module and layer-by-layer throughout the entire MLLM pipeline, enabling explicit visualization of where and how classification signals are distorted, diluted, or overridden.
As the first attempt to dissect classification performance degradation in medical MLLMs, our findings reveal four failure modes: 1) quality limitation in visual representation, 2) fidelity loss in connector projection, 3) comprehension deficit in LLM reasoning, and 4) misalignment of semantic mapping.
Meanwhile, we introduce quantitative scores that characterize the healthiness of feature evolution, enabling principled comparisons across diverse MLLMs and datasets. 
Furthermore, we provide insightful discussions centered on the critical barriers that prevent current medical MLLMs from fulfilling their promised clinical potential.
We hope that our work provokes rethinking within the community—highlighting that the road from high expectations to clinically deployable MLLMs remains long and winding.
\end{abstract}

\maketitle

\begin{figure}[t]
\centerline{\includegraphics[width=\linewidth]{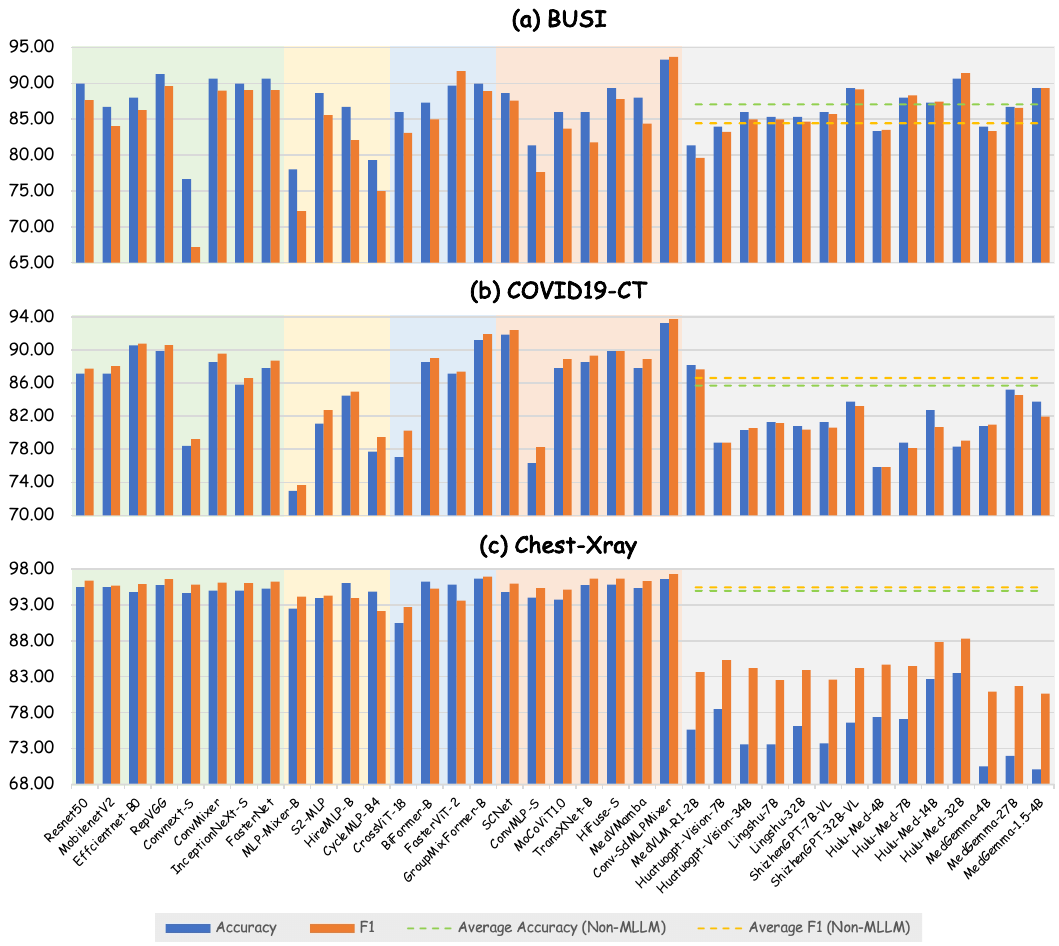}}
\caption{Performance comparison of state-of-the-art medical MLLMs and traditional deep learning methods across three medical image classification scenarios.
Model categories are distinguished by background color: CNN-based (green), MLP-based (yellow), ViT-based (blue), hybrid-based (red), and MLLM-based (gray).}
\label{f1}
\end{figure}

\section{Introduction}
The emergence of multimodal large language models (MLLMs) has marked a paradigm shift in artificial intelligence, seamlessly integrating visual understanding with linguistic reasoning~\citep{liu2023visual,li2023blip}. 
Leveraging the vast knowledge embedded in pre-trained language models, coupled with the capacity to interpret complex visual patterns through supervised fine-tuning (SFT), recent advances have extended MLLMs into the domain of medical imaging analysis.
In this field, medical MLLMs have demonstrated remarkable adaptability and competitive performance across diverse clinical tasks, ranging from medical visual question answering (VQA)~\citep{li2023llava}, radiology report generation~\citep{wu2025towards}, lesion detection~\citep{lu2024visual} to multi-task joint optimization~\citep{zhu2024uni}.
These achievements have ignited unprecedented enthusiasm for their potential to revolutionize clinical workflows, bringing the vision of democratizing access to expert-level diagnostic capabilities tantalizingly within reach.

As one of the earliest and most fundamental tasks integrated into the MLLM paradigm, medical image classification sits at the core of computational diagnosis, serving as the prerequisite for lesion detection, disease staging, and treatment planning~\citep{nam2025multimodal}.
Historically, traditional deep learning models such as convolutional neural networks (CNNs), multi-layer perceptrons (MLPs) and vision transformers (ViTs) have established robust baselines on specialized datasets.
Naturally, researchers have developed \textbf{h}igh-\textbf{y}ield \textbf{p}erformance \textbf{e}xpectations (\textbf{hype}) for medical MLLMs: the belief that massive pre-training data and billions of parameters should inherently yield superior discriminative power.
However, emerging evidence and discussions reveal an unsettling discrepancy that medical MLLMs consistently underperform compared to traditional deep learning methods.
The performance gap has been observed across diverse classification scenarios, including thoracic pathology diagnosis~\citep{fisher2025vision}, knee osteoarthritis grading~\citep{wang2026evaluating}, intracranial hemorrhage subtyping~\citep{wang2025zero}, skin cancer classification and respiratory disease diagnosis~\citep{raval2026llm}.
Our preliminary experiments further corroborate these findings.
As illustrated in Figure~\ref{f1}, current state-of-the-art medical MLLMs struggle to match the average performance of traditional deep learning methods.
Crucially, this degradation persists across medical imaging modalities, suggesting a systematic architectural limitation rather than dataset-specific anomalies.

The paradox mentioned above calls for a systematic investigation into the genesis and progression of performance degradation.
In the general domain, Zhang et al.~\citep{zhang2024visually} attribute the deficiency to the scarcity of fine-grained alignment data during multimodal pretraining, while Fu et al.~\citep{fu2025hidden} diagnose brittleness in task prompts and underutilization of visual representations by the LLM.
In the medical domain, Jeong et al.~\citep{jeong2024medical} question the efficacy of medical adaptation.
However, these pioneering works fall short of systematically dissecting how failures originate, propagate, and are either mitigated or exacerbated as visual information traverses the MLLM pipeline.
Without a fine-grained, layer-wise diagnosis that traces information flow across each module, efforts to improve the classification capability of medical MLLMs risk treating symptoms rather than root causes.

In this paper, we present the first systematic dissection of classification performance degradation in medical MLLMs.
Through extensive experiments on 14 open-source medical MLLMs across three representative classification tasks—ultrasound breast cancer classification, ‌computed tomography‌ (CT) COVID-19 diagnosis, and chest X-ray pneumonia screening—we move beyond superficial output-level benchmarking to diagnose the root causes of failure.
Specifically, we design a suite of probing experiments that track the information flow of visual features module-by-module and layer-by-layer throughout the entire MLLM pipeline, from the vision tower, through the connector, to the LLM and its final semantic output. This enables us to explicitly visualize where and how classification signals are distorted, diluted, or overridden.
Our diagnostic journey reveals four critical failure modes:
1) quality limitation in visual representation;
2) fidelity loss in connector projection;
3) comprehension deficit in LLM reasoning; 
and 4) misalignment of semantic mapping.
To enable quantitative comparisons, we introduce well-crafted scores that characterize the healthiness of feature evolution throughout the MLLM pipeline.
This quantitative lens enables objective cross-model comparisons of information preservation or loss across diverse modules and datasets, transforming subjective performance variations into measurable diagnostic criteria.
Beyond these diagnostic findings, we provide insightful discussions on three counter‑intuitive barriers that critically hinder medical MLLMs including the limited impact of medical adaptation, the suboptimal state of vision tower, and the comprehension-utilization entanglement in LLM reasoning.
We hope that our discussions can redirect the field's focus from the scale-centric hype of amassing data and parameters toward targeted innovations in architecture and mechanism design, thereby unlocking the transformative potential of MLLMs in clinical medicine.
In a nutshell, the contributions of this paper can be summarized as
follows:

\textbullet\; A systematic diagnosis. We conduct the first layer-wise, module-level dissection of classification performance degradation in medical MLLMs, revealing exactly where and how visual information is distorted throughout the pipeline.

\textbullet\; A taxonomy of failure modes. We identify and distill our observations into four critical failure modes, providing a unified vocabulary for diagnosing medical MLLMs.

\textbullet \; Apples-to-apples comparisons. We design novel scores that characterize the healthiness of feature evolution, offering a unified standard for fair cross-model evaluation.

\textbullet \; Insightful discussions. We reveal three intriguing barriers, with the hope of redirecting the field’s focus from scaling toward architectural and mechanistic innovations that unlock clinical potential.

\begin{figure*}[t]
\centerline{\includegraphics[width=\linewidth]{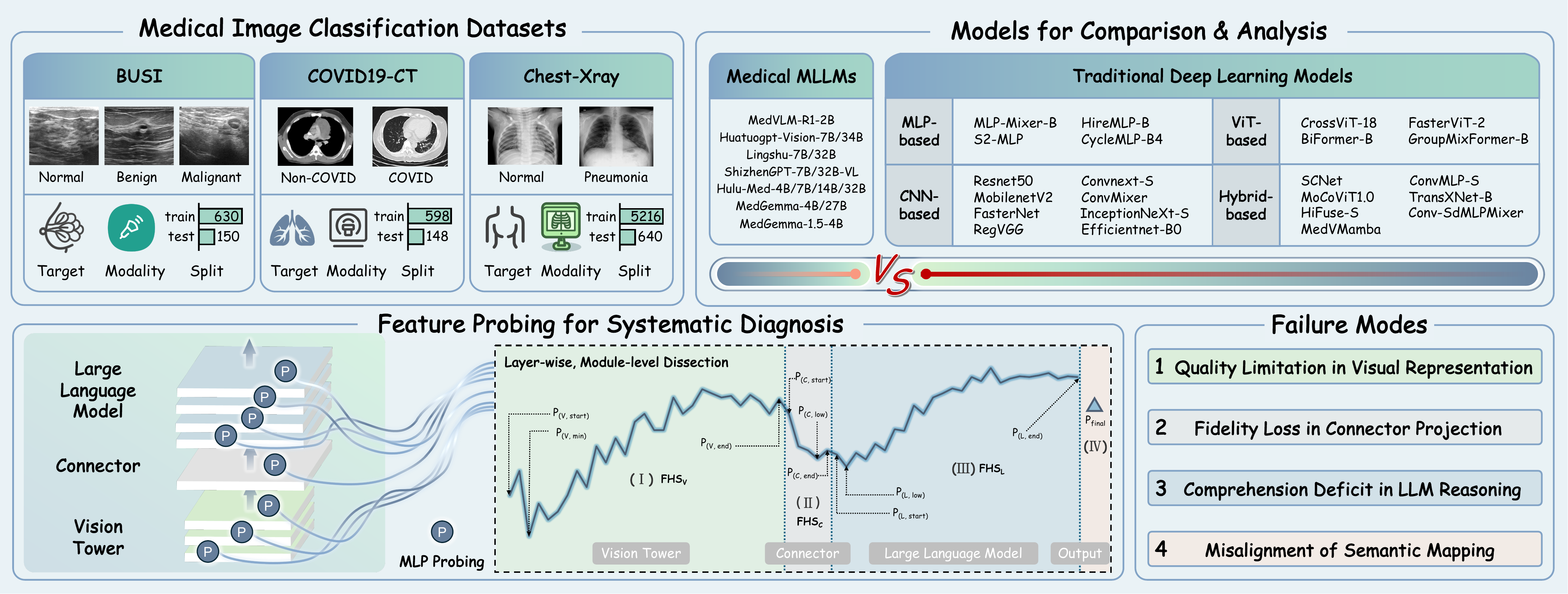}}
\caption{Evaluation framework for dissecting classification degradation in medical MLLMs.}
\label{f2}
\end{figure*}

\section{Related Work}
\subsection{Medical MLLM}
MLLMs typically consist of three core components~\citep{zhu2025connector}: a vision tower for raw visual feature extraction; a connector for cross-modal alignment; and an LLM backbone for auto-regressive text generation.
This architecture enables MLLMs to function as a general-purpose interface for diverse visual and linguistic tasks.
The success of general-domain MLLMs has spurred a surge of interest in adapting them to the medical domain~\citep{ye2025multimodal}, broadly following two development routes.
The first route assembles off-the-shelf vision towers and LLMs with custom-designed connectors, offering architectural flexibility for specific medical scenarios.
For example, XrayGPT~\citep{thawakar2024xraygpt} is designed for chest X-ray analysis, combining frozen MedClip~\citep{wang2022medclip} with Vicuna~\citep{chiang2023vicuna} via a linear projection connector.
Uni-Med~\citep{zhu2024uni} employs a connector mixture-of-experts to bridge EVA-CLIP~\citep{fang2023eva} and LLaMA2~\citep{touvron2023llama} for multi-task joint learning.
The second route directly fine-tunes general MLLMs on medical data, capitalizing on the rapid iteration and ever-strengthening capabilities of foundation models.
For instance, LLaVA-Med~\citep{li2023llava} adapts LLaVA through curriculum learning for medical VQA.
Several models built upon the Qwen-VL series~\citep{wang2024qwen2}-such as HuatuoGPT-Vision~\citep{chen2024towards}, Lingshu~\citep{xu2025lingshu}, and ShizhenGPT-VL~\citep{chen2025shizhengpt}—target distinct medical capabilities, ranging from medical multimodal understanding and reasoning to Traditional Chinese Medicine diagnostics.
While these developments signal exciting progress toward generalist medical AI, the field has been surprisingly silent on a critical question: given their overwhelming advantages in pre-training data and model parameters, do medical MLLMs actually deliver tangible gains over traditional methods on foundational tasks? The answer, as we will show, is far from straightforward.

\subsection{Medical Image Classification}
Medical image classification serves as a fundamental capability in medical image interpretation and computational diagnosis.
Over the past decade, four types of deep learning architectures have been developed to address this task.
CNN-based methods~\citep{he2016deep, tan2019efficientnet, chen2023run, yu2024inceptionnext} have been widely adopted due to their strong local feature extraction and parameter efficiency.
MLP-based methods~\citep{tolstikhin2021mlp,yu2022s2} replace convolutions with pure multi-layer perceptrons, achieving global receptive fields at the cost of local inductive bias.
ViT-based methods~\citep{chen2021crossvit, hatamizadeh2023fastervit} leverage self-attention to capture long-range dependencies, yet require substantial data for learning.
Hybrid-based methods~\citep{li2023convmlp,huo2024hifuse, ren2025conv} aim to integrate the advantages of different network architectures by balancing local detail and global context.
Rather than employing dedicated classification heads, medical MLLMs reformulate image classification as language-based VQA.
By leveraging the reasoning and instruction-following capabilities of the LLM, classification results are generated either as a choice among predefined options~\citep{singh2025crossmed} or as an open-ended response directly specifying the category~\citep{zhu2025enhancing}.
Despite the intuitive appeal of handling both classification and more complex medical tasks within the unified framework, emerging observations reveal that medical MLLMs consistently underperform compared to traditional deep learning methods across diverse classification tasks~\citep{wang2025zero,wang2026evaluating,raval2026llm}.
While these studies provide valuable discussions, they largely remain at the phenomenological level.
Our work provides the first systematic diagnosis of classification performance degradation in medical MLLMs, revealing four failure modes and three critical barriers that previous studies have overlooked.

\subsection{Feature Probing}
By training lightweight classifiers on frozen intermediate activations, probing reveals which information is encoded at specific network depths and serves as a diagnostic tool for representation quality~\citep{alain2016understanding, conneau2018you}.
Probing originates in natural language processing, where it reveals that language models encode hierarchical linguistic information in a layer-wise fashion~\citep{van2019does, hewitt2019designing}.
In computer vision, probing demonstrates a layer-wise progression from edges and textures to object parts and holistic concepts~\citep{raghu2021vision}.
With the advent of multimodal models, probing has been extended to analyze cross-modal representations.
Several studies have investigated whether visual concepts survive projection into the language space~\citep{verma2024cross} and how multimodal fusion affects representational quality~\citep{wang2025cross}.
Zhang et al.~\citep{zhang2024visually} employ linear probes to show that classification-critical information persists in vision tower outputs but remains inaccessible to the LLM.
Despite these advances, existing probing studies in medical MLLMs typically focus on isolated or specific layers, driven by individual study objectives.
This fragmentation prevents a holistic understanding of how classification signals transform across the pipeline, while quantitative metrics to characterize such evolutionary dynamics remain absent.
In this work, we conduct a pipeline-wide, layer-wise probing analysis in diverse medical MLLMs.
Within each module, we further introduce quantitative metrics to characterize the healthiness of feature evolution, enabling principled comparisons of different models and components.

\begin{figure*}[ht]
\centerline{\includegraphics[width=0.957\linewidth]{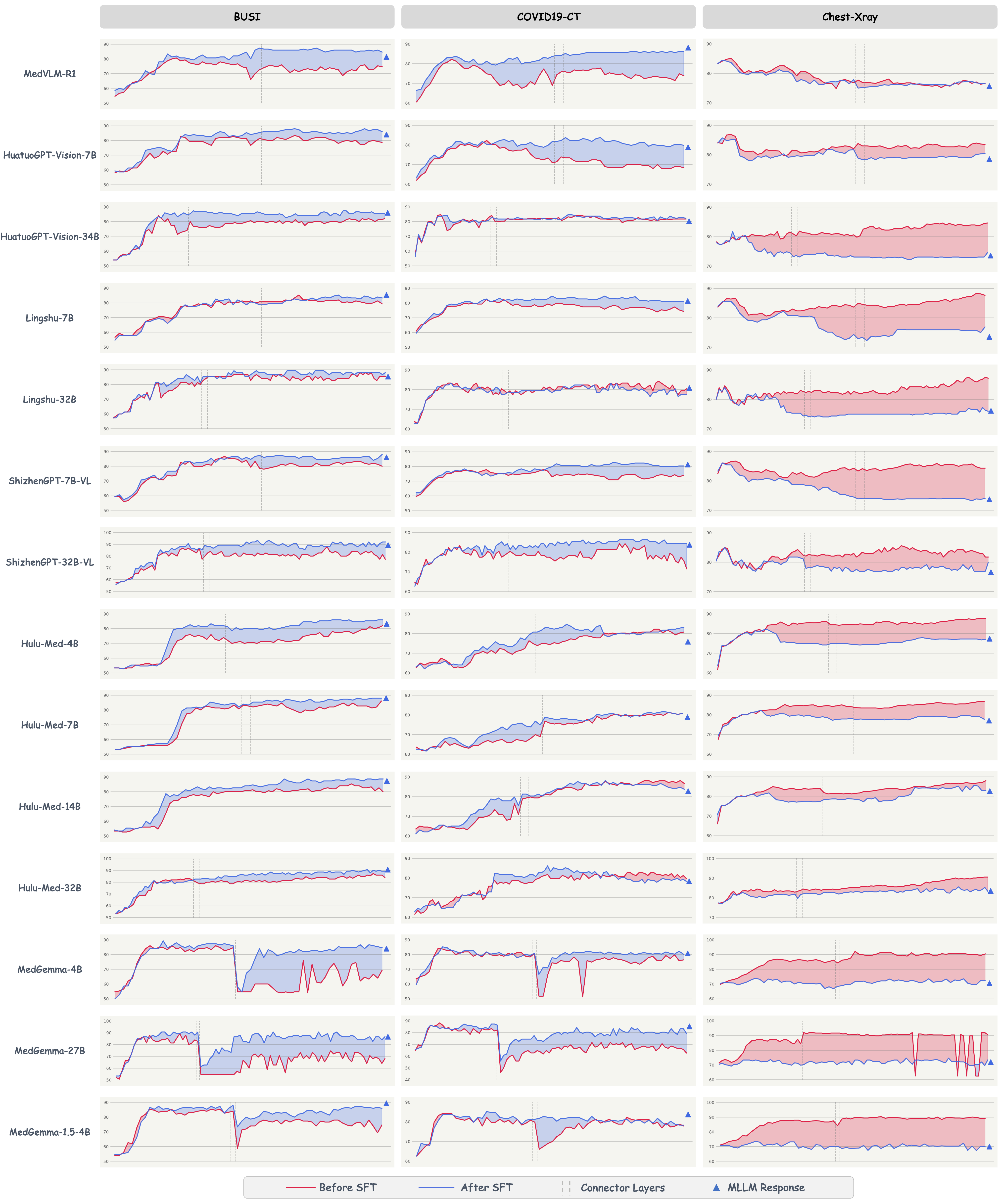}}
\caption{Layer-wise probing accuracy curves for 14 medical MLLMs across three datasets.}
\label{f3}
\end{figure*}

\section{Evaluation Protocol}
As illustrated in Figure~\ref{f2}, this section presents the evaluation setup for systematically dissecting classification performance of medical MLLMs, including the selected medical image classification datasets, the models for comparison and analysis, the implementation details of feature probing, and the evaluation metrics.

\subsection{Datasets}
\paragraph{Ultrasound for Breast Cancer Classification}
The Breast Ultrasound Images (BUSI) dataset~\citep{al2020dataset} is collected from 600 female patients at Baheya Hospital, Cairo, Egypt.
It contains 780 breast ultrasound images with an average resolution of 500*500 pixels.
BUSI is divided into three categories: normal, benign, and malignant, containing 133, 437, and 210 images respectively.

\paragraph{CT for COVID-19 Diagnosis}
The COVID19-CT dataset~\citep{yang2020covid}  is constructed by extracting CT images from 760 COVID-19 preprints on medRxiv and bioRxiv.
It comprises 349 COVID-19 positive CT images from 216 patients and 397 negative samples.

\paragraph{Xray for Pneumonia Screening}
The Chest-Xray dataset~\citep{kermany2018identifying} is derived from a retrospective cohort of pediatric patients aged one to five years at Guangzhou Women and Children's Medical Center.
It contains 5,856 chest X-ray images, labeled as normal (1,583 images) and pneumonia (4,273 images).

Detailed instruction templates and examples for each classification task are provided in Supplementary Materials A.

\subsection{Models and Implementation Details}
We follow the comparison protocol of Ren et al.~\citep{ren2025conv} and include four types of traditional deep learning models: 8 CNN-based, 4 MLP-based, 4 ViT-based, and 7 hybrid-based methods.
For medical MLLMs, we select 14 open-source models spanning six representative series:
MedVLM-R1~\citep{pan2025medvlm},
HuatuoGPT-Vision-7B/34B~\citep{chen2024towards},
Lingshu-7B/32B~\citep{xu2025lingshu},
ShizhenGPT-7B/32B-VL~\citep{chen2025shizhengpt},
Hulu-Med-4B/7B
/14B/32B~\citep{jiang2025hulu},
MedGemma-4B/27B~\citep{sellergren2025medgemma},
and MedGemma-1.5-4B.
Detailed architecture designs, training stages, and data sources for each medical MLLM are provided in Supplementary Materials B.

Our main experiments consist of three complementary parts to systematically diagnose classification performance in MLLMs: 
(1) Dataset-specific SFT of each medical MLLM.
We apply LoRA~\citep{hu2021lora} to all linear layers with rank 8 and alpha 32, efficiently adapting the model to the target classification task.
(2) Feature probing on the original medical MLLMs. 
For each module—vision tower, connector, and LLM—and for each layer within these modules, we attach a lightweight probing head consisting of a two-layer MLP with ReLU activation and dropout rate 0.1.
The probe parameters are optimized while the MLLM backbone remains frozen, ensuring that the probe accuracy reflects the amount of task-relevant information preserved at that specific point in the pipeline.
(3) Feature probing on fine-tuned medical MLLMs.
We repeat the same probing procedure to examine how supervised adaptation alters the information flow and whether it strengthens or weakens representational capabilities.
We use the AdamW optimizer with a learning rate of 1e-4.
A cosine learning rate scheduler with a warmup ratio of 0.05 is adopted.
All experiments are trained twice for 20 epochs with a batch size of 4, separately using 8 NVIDIA A800 GPUs and 8 Ascend 910B NPUs.

\subsection{Metrics}
We employ standard classification metrics including accuracy, F1-score, precision, recall, and AUC to  evaluate model and probe performance.
Drawing inspiration from the Sharpe Ratio in financial economics, which evaluates return relative to risk, we propose the feature health score (FHS) to holistically assess how well a module preserves and refines task-relevant information across its layers.
Formally, let \(V\), \(C\), and \(L\) represent the vision tower, connector, and large language model, respectively.
For a given module \(M \in \{V, C, L\}\) with $n$ layers, we introduce 
growth factor ($GF_M$) and volatility penalty ($VP_M$), defined as follows:
\begin{equation}
GF_M=\ln \left(P_{(M, \text{end})} / {\sqrt{P_{(M, \text{start})} P_{(M, \text{min})}}}\right) 
\end{equation}
\begin{equation}
VP_M=\exp \left(-\lambda \cdot \sum_{i=1}^{n-1}\left|P_{\left(M,i+1\right)}-P_{\left(M,i\right)}\right| / {P_{(M, \text{end})}}\right) 
\end{equation}
where \(P_{(M,\text{start})}\), \(P_{(M,\text{min})}\), and \(P_{(M,\text{end})}\) represent the probing performance at the first, worst, and last layer of module \(M\), respectively.
$i$ is the layer index.
$\lambda$ is set to 0.2 to control the strength of penalty.
The former measures the net improvement from the module’s early and worst layers to its final output.
The latter quantifies the stability of the feature evolution path.
High total variation indicates erratic oscillations, sharp peaks, or sudden drops, which are signs of architectural fragility or non‑robust features.
The FHS of module $M$ can be formed as follows:
\begin{equation}
FHS_M=P_{(M, \text{end})}  \cdot \left(1+GF_M\right) \cdot VP_M
\end{equation}
which ensures that a module receives a high FHS only if it simultaneously achieves strong final performance, demonstrates meaningful improvement over its early and worst layers, and maintains a smooth, stable trajectory.
Therefore, we can obtain a four‑score profile:
(\MakeUppercase{\romannumeral 1}) \(FHS_V\), (\MakeUppercase{\romannumeral 2}) \(FHS_C\), (\MakeUppercase{\romannumeral 3}) \(FHS_L\),
and (\MakeUppercase{\romannumeral 4}) \(P_{final}\), the probing performance at the final semantic output of the MLLM.
The evolution process of feature healthiness offers a holistic and quantitative view, enabling principled apples-to-apples comparisons across different MLLMs.

\section{Evaluation Results}
This section presents the evaluation analysis of classification performance degradation inside medical MLLMs.
We first provide an overview of our layer-wise, module-level probing results, then distill the observed patterns into four failure modes, and finally apply the proposed quantitative metrics to enable principled cross-model comparisons.
Figure~\ref{f3} visualizes the probing accuracy curves for all 14 medical MLLMs across the three datasets before and after task-specific fine-tuning.
Additional probing metrics are provided in Supplementary Material C.

\subsection{Initial Observations}
To minimize cognitive interference from the diverse pre-training data, model architectures, and parameter scales across different medical MLLMs, we first focus on within-module pattern analysis.
Specifically, we examine the 84 probing accuracy curves and summarize the recurring behaviors observed in the vision tower, connector, LLM, and final semantic output, respectively.

Within the vision tower, the probing accuracy curves exhibit considerable variation across datasets and models.
On the BUSI and COVID-19 CT datasets, the most common pattern is a rise-and-fluctuate trend: the accuracy rises steadily from early layers, reaches a peak in mid-to-deep layers, and then fluctuates within a narrow range.
The curves before and after fine-tuning show similar shapes, indicating that task-specific adaptation does not fundamentally alter how the encoder organizes visual information.
However, the behavior diverges markedly on the Chest-Xray dataset.
On the one hand, the probing accuracy may exhibit a sustained decline as layers deepen.
On the other hand, the curves before and after fine-tuning differ in shape.
This counterintuitive observation suggests that dataset-specific fine-tuning may disrupt rather than enhance the discriminative capacity of the vision tower, leading to cumulative degradation with depth.

Probing accuracy within the connector exhibits a distinct pattern: it largely preserves the discriminative capability of the vision tower output layer, with only negligible upward or downward fluctuations.
Such limited variation suggests that the connector primarily performs dimensional alignment between the vision tower's output space and the LLM's input embedding space, rather than functioning as a sophisticated feature refiner.
In other words, the connector does not substantially enhance or impair the discriminative information relevant to classification.

The LLM exhibits the most diverse behaviors among the three modules.
First, probing accuracy may drop sharply at the first LLM layer then recovers steadily, indicating that the LLM initially struggles to decode visual information but gradually regains discriminability.
Second, the curve remains largely unchanged, suggesting that the LLM merely inherits the representation quality without further refinement.
Third, accuracy gradually decreases with depth.
This occurs when upstream modules already suffer from degraded representations, and the LLM fails to salvage them.

We observe that the final semantic output accuracy deviates from the probing accuracy at the last LLM layer: sometimes lower, occasionally higher, but rarely equal.
Discriminative representations do not guarantee a correct generative semantic output, suggesting that the generative decoding process may introduce an additional gap independent of representation quality.

\subsection{Failure Modes}
Synthesizing the observations above, we distill the recurring degradation patterns into four critical failure modes that collectively explain why medical MLLMs underperform in image classification.

\paragraph{Quality Limitation in Visual Representation.}
The vision tower fails to encode task-specific discriminative features for fine-grained medical patterns, producing generic representations that lack diagnostic specificity.
The consequence is an irrecoverable information deficit at the pipeline's inception: downstream modules cannot reconstruct critical visual cues (e.g., lesion boundaries, tissue textures) from impoverished inputs, establishing a performance ceiling that no subsequent processing can breach.

\paragraph{Fidelity Loss in Connector Projection.}
While designed to bridge modality gaps, the connector often operates as a lossy compression channel, diluting fine-grained spatial and textural details essential for medical diagnosis in favor of coarse semantic alignment.
Critical diagnostic evidence is silently eroded during projection, leaving the LLM with ambiguous or incomplete visual evidence despite technically aligned representations.

\paragraph{Comprehension Deficit in LLM Reasoning.}
This failure mode arises from the causal attention mechanism inherent to autoregressive LLMs.
Specifically, the LLM's intermediate representations of image tokens evolve solely through self-attention within the visual prefix, without ever being contextualized by the task-relevant language that follows.
Consequently, the LLM fails to interpret visual evidence in a medically meaningful way, yielding little discriminative gain, or even incurring degradation.

\paragraph{Misalignment of Semantic Mapping.}
A fundamental incompatibility exists between the generative text decoding objective and the discriminative nature of classification.
The LLM is trained to produce fluent, open-ended text, not to map latent representations to a fixed set of categorical outputs with high fidelity.
This misalignment manifests as a persistent gap between representation quality and output correctness.

\begin{table*}[ht]
\small
\setlength{\tabcolsep}{2pt}
\caption{Feature health scores ($\bm{FHS_V} \rightarrow \bm{FHS_C} \rightarrow \bm{FHS_L} \rightarrow \bm{P_{\text{final}}}$) for 14 medical MLLMs across three classification datasets. Results in \textbf{bold} denote the dataset best, while results with \underline{underlines} indicate the dataset worst.}
\label{t1}
\begin{tabular}{lcccccccp{0.3cm}cccccccp{0.3cm}ccccccc}
\toprule
\textbf{Medical MLLMs} & \multicolumn{7}{c}{\textbf{BUSI}} & & \multicolumn{7}{c}{\textbf{COVID19-CT}} & & \multicolumn{7}{c}{\textbf{Chest-Xray}} \\
\midrule
\cellcolor[HTML]{F2F2F2}MedVLM-R1-2B & 
\cellcolor[HTML]{F2F2F2}\underline{91.42} & \cellcolor[HTML]{F2F2F2}$\rightarrow$ & \cellcolor[HTML]{F2F2F2}88.41 & \cellcolor[HTML]{F2F2F2}$\rightarrow$ & \cellcolor[HTML]{F2F2F2}81.21 & \cellcolor[HTML]{F2F2F2}$\rightarrow$ & \cellcolor[HTML]{F2F2F2}\underline{81.33} & 
\cellcolor[HTML]{F2F2F2} & 
\cellcolor[HTML]{F2F2F2}93.46 & \cellcolor[HTML]{F2F2F2}$\rightarrow$ & \cellcolor[HTML]{F2F2F2}84.24 & \cellcolor[HTML]{F2F2F2}$\rightarrow$ & \cellcolor[HTML]{F2F2F2}\textbf{86.11} & \cellcolor[HTML]{F2F2F2}$\rightarrow$ & \cellcolor[HTML]{F2F2F2}\textbf{88.18} & 
\cellcolor[HTML]{F2F2F2} & 
\cellcolor[HTML]{F2F2F2}71.58 & \cellcolor[HTML]{F2F2F2}$\rightarrow$ & \cellcolor[HTML]{F2F2F2}75.29 & \cellcolor[HTML]{F2F2F2}$\rightarrow$ & \cellcolor[HTML]{F2F2F2}76.44 & \cellcolor[HTML]{F2F2F2}$\rightarrow$ & \cellcolor[HTML]{F2F2F2}75.64 \\
Huatuogpt-Vision-7B & 
101.88 & $\rightarrow$ & 85.87 & $\rightarrow$ & 83.25 & $\rightarrow$ & 84.00 & 
& 
92.21 & $\rightarrow$ & 82.66 & $\rightarrow$ & 74.73 & $\rightarrow$ & 78.82 & 
& 
78.10 & $\rightarrow$ & 78.03 & $\rightarrow$ & 81.35 & $\rightarrow$ & 78.53 \\
\cellcolor[HTML]{F2F2F2}Huatuogpt-Vision-34B & 
\cellcolor[HTML]{F2F2F2}109.59 & \cellcolor[HTML]{F2F2F2}$\rightarrow$ & \cellcolor[HTML]{F2F2F2}88.95 & \cellcolor[HTML]{F2F2F2}$\rightarrow$ & \cellcolor[HTML]{F2F2F2}\underline{79.56} & \cellcolor[HTML]{F2F2F2}$\rightarrow$ & \cellcolor[HTML]{F2F2F2}86.00 & 
\cellcolor[HTML]{F2F2F2} & 
\cellcolor[HTML]{F2F2F2}96.34 & \cellcolor[HTML]{F2F2F2}$\rightarrow$ & \cellcolor[HTML]{F2F2F2}81.92 & \cellcolor[HTML]{F2F2F2}$\rightarrow$ & \cellcolor[HTML]{F2F2F2}77.58 & \cellcolor[HTML]{F2F2F2}$\rightarrow$ & \cellcolor[HTML]{F2F2F2}80.30 & 
\cellcolor[HTML]{F2F2F2} & 
\cellcolor[HTML]{F2F2F2}69.02 & \cellcolor[HTML]{F2F2F2}$\rightarrow$ & \cellcolor[HTML]{F2F2F2}73.81 & \cellcolor[HTML]{F2F2F2}$\rightarrow$ & \cellcolor[HTML]{F2F2F2}74.15 & \cellcolor[HTML]{F2F2F2}$\rightarrow$ & \cellcolor[HTML]{F2F2F2}73.56 \\
Lingshu-7B & 
97.69 & $\rightarrow$ & 77.74 & $\rightarrow$ & 83.34 & $\rightarrow$ & 85.33 & 
& 
\textbf{103.75} & $\rightarrow$ & 83.25 & $\rightarrow$ & 76.81 & $\rightarrow$ & 81.28 & 
& 
66.30 & $\rightarrow$ & 73.78 & $\rightarrow$ & 80.00 & $\rightarrow$ & 73.56 \\
\cellcolor[HTML]{F2F2F2}Lingshu-32B & 
\cellcolor[HTML]{F2F2F2}95.03 & \cellcolor[HTML]{F2F2F2}$\rightarrow$ & \cellcolor[HTML]{F2F2F2}85.33 & \cellcolor[HTML]{F2F2F2}$\rightarrow$ & \cellcolor[HTML]{F2F2F2}81.95 & \cellcolor[HTML]{F2F2F2}$\rightarrow$ & \cellcolor[HTML]{F2F2F2}85.33 & 
\cellcolor[HTML]{F2F2F2} & 
\cellcolor[HTML]{F2F2F2}84.52 & \cellcolor[HTML]{F2F2F2}$\rightarrow$ & \cellcolor[HTML]{F2F2F2}79.22 & \cellcolor[HTML]{F2F2F2}$\rightarrow$ & \cellcolor[HTML]{F2F2F2}\underline{66.71} & \cellcolor[HTML]{F2F2F2}$\rightarrow$ & \cellcolor[HTML]{F2F2F2}80.79 & 
\cellcolor[HTML]{F2F2F2} & 
\cellcolor[HTML]{F2F2F2}65.90 & \cellcolor[HTML]{F2F2F2}$\rightarrow$ & \cellcolor[HTML]{F2F2F2}73.59 & \cellcolor[HTML]{F2F2F2}$\rightarrow$ & \cellcolor[HTML]{F2F2F2}75.61 & \cellcolor[HTML]{F2F2F2}$\rightarrow$ & \cellcolor[HTML]{F2F2F2}76.12 \\
ShizhenGPT-7B-VL & 
102.14 & $\rightarrow$ & 87.87 & $\rightarrow$ & 87.94 & $\rightarrow$ & 86.00 & 
& 
90.02 & $\rightarrow$ & 80.10 & $\rightarrow$ & 78.06 & $\rightarrow$ & 81.28 & 
& 
66.39 & $\rightarrow$ & 74.17 & $\rightarrow$ & 73.80 & $\rightarrow$ & 73.72 \\
\cellcolor[HTML]{F2F2F2}ShizhenGPT-32B-VL & 
\cellcolor[HTML]{F2F2F2}105.04 & \cellcolor[HTML]{F2F2F2}$\rightarrow$ & \cellcolor[HTML]{F2F2F2}87.07 & \cellcolor[HTML]{F2F2F2}$\rightarrow$ & \cellcolor[HTML]{F2F2F2}82.18 & \cellcolor[HTML]{F2F2F2}$\rightarrow$ & \cellcolor[HTML]{F2F2F2}89.33 & 
\cellcolor[HTML]{F2F2F2} & 
\cellcolor[HTML]{F2F2F2}99.61 & \cellcolor[HTML]{F2F2F2}$\rightarrow$ & \cellcolor[HTML]{F2F2F2}85.10 & \cellcolor[HTML]{F2F2F2}$\rightarrow$ & \cellcolor[HTML]{F2F2F2}79.00 & \cellcolor[HTML]{F2F2F2}$\rightarrow$ & \cellcolor[HTML]{F2F2F2}83.74 & 
\cellcolor[HTML]{F2F2F2} & 
\cellcolor[HTML]{F2F2F2}78.55 & \cellcolor[HTML]{F2F2F2}$\rightarrow$ & \cellcolor[HTML]{F2F2F2}78.46 & \cellcolor[HTML]{F2F2F2}$\rightarrow$ & \cellcolor[HTML]{F2F2F2}75.81 & \cellcolor[HTML]{F2F2F2}$\rightarrow$ & \cellcolor[HTML]{F2F2F2}76.60 \\
Hulu-Med-4B & 
107.01 & $\rightarrow$ & \underline{76.83} & $\rightarrow$ & 88.58 & $\rightarrow$ & 83.33 & 
& 
84.92 & $\rightarrow$ & 81.98 & $\rightarrow$ & 81.36 & $\rightarrow$ & \underline{75.86} & 
& 
78.66 & $\rightarrow$ & 74.84 & $\rightarrow$ & 78.26 & $\rightarrow$ & 77.40 \\
\cellcolor[HTML]{F2F2F2}Hulu-Med-7B & 
\cellcolor[HTML]{F2F2F2}106.55 & \cellcolor[HTML]{F2F2F2}$\rightarrow$ & \cellcolor[HTML]{F2F2F2}82.87 & \cellcolor[HTML]{F2F2F2}$\rightarrow$ & \cellcolor[HTML]{F2F2F2}86.94 & \cellcolor[HTML]{F2F2F2}$\rightarrow$ & \cellcolor[HTML]{F2F2F2}88.00 & 
\cellcolor[HTML]{F2F2F2} & 
\cellcolor[HTML]{F2F2F2}80.78 & \cellcolor[HTML]{F2F2F2}$\rightarrow$ & \cellcolor[HTML]{F2F2F2}\underline{77.98} & \cellcolor[HTML]{F2F2F2}$\rightarrow$ & \cellcolor[HTML]{F2F2F2}80.69 & \cellcolor[HTML]{F2F2F2}$\rightarrow$ & \cellcolor[HTML]{F2F2F2}78.82 & 
\cellcolor[HTML]{F2F2F2} & 
\cellcolor[HTML]{F2F2F2}81.37 & \cellcolor[HTML]{F2F2F2}$\rightarrow$ & \cellcolor[HTML]{F2F2F2}78.17 & \cellcolor[HTML]{F2F2F2}$\rightarrow$ & \cellcolor[HTML]{F2F2F2}76.42 & \cellcolor[HTML]{F2F2F2}$\rightarrow$ & \cellcolor[HTML]{F2F2F2}77.08 \\
Hulu-Med-14B & 
105.23 & $\rightarrow$ & 81.54 & $\rightarrow$ & 90.29 & $\rightarrow$ & 87.33 & 
& 
84.00 & $\rightarrow$ & 81.28 & $\rightarrow$ & 81.41 & $\rightarrow$ & 82.76 & 
& 
79.95 & $\rightarrow$ & 78.65 & $\rightarrow$ & \textbf{81.89} & $\rightarrow$ & 82.69 \\
\cellcolor[HTML]{F2F2F2}Hulu-Med-32B & 
\cellcolor[HTML]{F2F2F2}103.12 & \cellcolor[HTML]{F2F2F2}$\rightarrow$ & \cellcolor[HTML]{F2F2F2}82.67 & \cellcolor[HTML]{F2F2F2}$\rightarrow$ & \cellcolor[HTML]{F2F2F2}86.20 & \cellcolor[HTML]{F2F2F2}$\rightarrow$ & \cellcolor[HTML]{F2F2F2}\textbf{90.67} & 
\cellcolor[HTML]{F2F2F2} & 
\cellcolor[HTML]{F2F2F2}\underline{78.46} & \cellcolor[HTML]{F2F2F2}$\rightarrow$ & \cellcolor[HTML]{F2F2F2}82.27 & \cellcolor[HTML]{F2F2F2}$\rightarrow$ & \cellcolor[HTML]{F2F2F2}67.76 & \cellcolor[HTML]{F2F2F2}$\rightarrow$ & \cellcolor[HTML]{F2F2F2}78.33 & 
\cellcolor[HTML]{F2F2F2} & 
\cellcolor[HTML]{F2F2F2}\textbf{83.56} & \cellcolor[HTML]{F2F2F2}$\rightarrow$ & \cellcolor[HTML]{F2F2F2}\textbf{83.41} & \cellcolor[HTML]{F2F2F2}$\rightarrow$ & \cellcolor[HTML]{F2F2F2}77.91 & \cellcolor[HTML]{F2F2F2}$\rightarrow$ & \cellcolor[HTML]{F2F2F2}\textbf{83.49} \\
MedGemma-4B & 
\textbf{117.15} & $\rightarrow$ & 86.00 & $\rightarrow$ & \textbf{100.87} & $\rightarrow$ & 84.00 & 
& 
89.83 & $\rightarrow$ & 80.79 & $\rightarrow$ & 86.08 & $\rightarrow$ & 80.79 & 
& 
\underline{59.91} & $\rightarrow$ & \underline{68.43} & $\rightarrow$ & 68.96 & $\rightarrow$ & 70.51 \\
\cellcolor[HTML]{F2F2F2}MedGemma-27B & 
\cellcolor[HTML]{F2F2F2}115.58 & \cellcolor[HTML]{F2F2F2}$\rightarrow$ & \cellcolor[HTML]{F2F2F2}\textbf{90.67} & \cellcolor[HTML]{F2F2F2}$\rightarrow$ & \cellcolor[HTML]{F2F2F2}83.91 & \cellcolor[HTML]{F2F2F2}$\rightarrow$ & \cellcolor[HTML]{F2F2F2}86.67 & 
\cellcolor[HTML]{F2F2F2} & 
\cellcolor[HTML]{F2F2F2}101.94 & \cellcolor[HTML]{F2F2F2}$\rightarrow$ & \cellcolor[HTML]{F2F2F2}\textbf{86.27} & \cellcolor[HTML]{F2F2F2}$\rightarrow$ & \cellcolor[HTML]{F2F2F2}79.80 & \cellcolor[HTML]{F2F2F2}$\rightarrow$ & \cellcolor[HTML]{F2F2F2}85.22 & 
\cellcolor[HTML]{F2F2F2} & 
\cellcolor[HTML]{F2F2F2}68.76 & \cellcolor[HTML]{F2F2F2}$\rightarrow$ & \cellcolor[HTML]{F2F2F2}69.87 & \cellcolor[HTML]{F2F2F2}$\rightarrow$ & \cellcolor[HTML]{F2F2F2}65.01 & \cellcolor[HTML]{F2F2F2}$\rightarrow$ & \cellcolor[HTML]{F2F2F2}71.96 \\
MedGemma-1.5-4B & 
109.54 & $\rightarrow$ & 88.00 & $\rightarrow$ & 90.14 & $\rightarrow$ & 89.33 & 
& 
91.73 & $\rightarrow$ & 81.28 & $\rightarrow$ & 70.44 & $\rightarrow$ & 83.74 & 
& 
64.13 & $\rightarrow$ & 70.19 & $\rightarrow$ & \underline{63.12} & $\rightarrow$ & \underline{70.03} \\

\cellcolor[HTML]{D9D9D9}Average & 
\cellcolor[HTML]{D9D9D9}104.78 & \cellcolor[HTML]{D9D9D9} & \cellcolor[HTML]{D9D9D9}84.99 & \cellcolor[HTML]{D9D9D9} & \cellcolor[HTML]{D9D9D9}86.17 & \cellcolor[HTML]{D9D9D9} & \cellcolor[HTML]{D9D9D9}86.19 & 
\cellcolor[HTML]{D9D9D9} & 
\cellcolor[HTML]{D9D9D9}90.83 & \cellcolor[HTML]{D9D9D9} & \cellcolor[HTML]{D9D9D9}82.02 & \cellcolor[HTML]{D9D9D9} & \cellcolor[HTML]{D9D9D9}77.61 & \cellcolor[HTML]{D9D9D9} & \cellcolor[HTML]{D9D9D9}81.42 & 
\cellcolor[HTML]{D9D9D9} & 
\cellcolor[HTML]{D9D9D9}72.30 & \cellcolor[HTML]{D9D9D9} & \cellcolor[HTML]{D9D9D9}75.05 & \cellcolor[HTML]{D9D9D9} & \cellcolor[HTML]{D9D9D9}74.91 & \cellcolor[HTML]{D9D9D9} & \cellcolor[HTML]{D9D9D9}75.78 \\

\bottomrule
\end{tabular}
\end{table*}

\subsection{Quantitative Comparisons}
Table~\ref{t1} presents the four-score profiles ($FHS_V$, $FHS_C$, $FHS_L$, $P_{final}$) for 14 medical MLLMs across three datasets.
This standardized representation transforms diverse architectural complexities into comparable trajectories of feature healthiness, enabling progressive comparisons from the perspective of modules, models, and datasets.

\paragraph{Module-level Analysis.}
By examining each score column vertically, we can identify which models excel or falter at specific stages of the pipeline.
The connector score \(FHS_C\) shows the least variation across models: the best and worst values are closer, indicating that the connector’s impact is secondary.
In contrast, the vision tower score \(FHS_V\) and LLM score \(FHS_L\) range widely between the best and worst models, highlighting that the quality limitation in visual representation and the comprehension deficit in LLM reasoning are the dominant bottlenecks.

\paragraph{Model-level Analysis.}
Achieving a high \(P_{\text{final}}\) necessitates robust health across all preceding modules, with no single stage exhibiting catastrophic failure.
Conversely, a poor \(P_{\text{final}}\) invariably correlates with at least one severely degraded module.
Furthermore, the four-score profile of each model challenges the assumption that parameter scaling alone ensures superior outcomes.
Within model series, increasing model size does not consistently improve all $FHS$ metrics, underscoring that larger models do not inherently resolve the identified failure modes.

\paragraph{Dataset-level Analysis.}
Examining the four-score profiles across datasets reveals clear hierarchical trends.
Averaged over all models, BUSI consistently achieves the highest scores, followed by COVID-19 CT, with Chest-Xray lagging substantially behind.
This descending trend mirrors the performance gap observed between medical MLLMs and traditional deep learning models.
Moreover, the dataset-specific fragility exposes a critical limitation of current generalist medical MLLMs: despite training on diverse imaging modalities and task formats, their capability remains highly dependent on the specific downstream distribution.

\begin{figure}[t]
\centerline{\includegraphics[width=\linewidth]{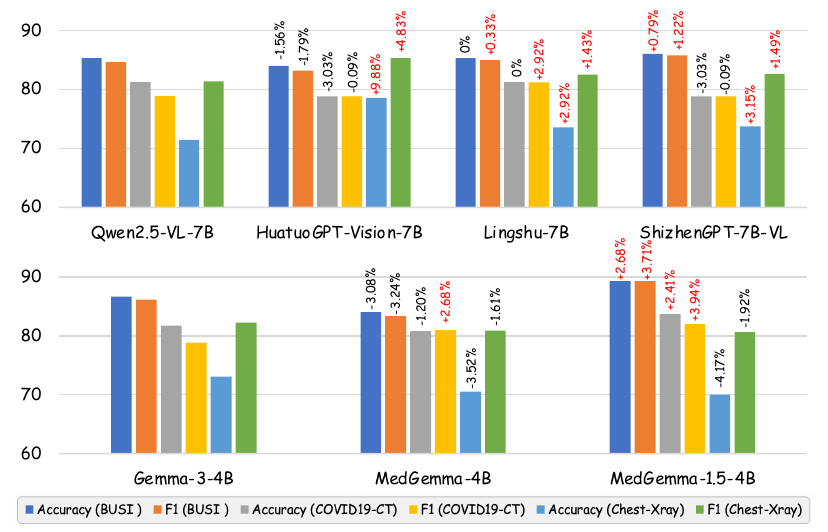}}
\caption{Marginal gains of medical adaptation over general-domain MLLMs: HuatuoGPT-Vision-7B/Lingshu-7B/ShizhenGPT-7B-VL vs. Qwen2.5-VL-7B; MedGemma-4B/MedGemma-1.5-4B vs. Gemma-3-4B.}
\label{f4}
\end{figure}

\section{Insightful Discussions}
Even when feature evolution within medical MLLMs appears healthy, the final classification accuracy may fall short of clinical expectations and underperform traditional deep learning models.
In this section, we dissect three critical barriers that hinder medical MLLMs in image classification.

\paragraph{The Limited Impact of Medical Adaptation.}
To validate whether extensive medical adaptation for MLLM yields tangible benefits for downstream medical image classification, we conduct two groups of experiments illustrated in Figure~\ref{f4}.
Specifically, we compare Qwen2.5-VL-7B and Gemma-3-4B directly fine-tuned on target datasets against the specialized medical MLLMs derived from them, respectively.
Across three datasets, the improvements from medical adaptation remain minimal.
For the Qwen-VL series, HuatuoGPT-Vision-7B, Lingshu-7B, and ShizhenGPT-7B-VL achieve average accuracy gains of only 1.37\%, 1.27\%, and 0.59\%, respectively.
For the Gemma series, MedGemma-4B actually underperforms its base model by 1.66\%, while MedGemma-1.5-4B yields a modest 1.17\% improvement.
Despite the substantial cost in terms of data and computational resources for domain‑specific pre‑training and fine‑tuning, the resulting medical MLLMs offer only negligible classification benefits over their general‑domain counterparts.

\paragraph{The Suboptimal State of Vision Tower.}
A more critical barrier lies in the vision tower itself.
As the core module that determines the quality of visual representation, the vision tower is rarely optimized in a way that directly benefits classification.
We first examine whether updating the vision tower during task-specific fine-tuning actually helps.
Table~\ref{t2} compares HuatuoGPT-Vision-7B and MedGemma-4B under two settings: with the vision tower frozen versus fine-tuned via LoRA.
The average gaps of -0.38\% and 0.27\% indicate that unfreezing the vision tower does not reliably improve classification and may even harm it.

We further compare three training paradigms using ShizhenGPT-7B-VL and MedGemma-1.5-4B: 
standard MLLM SFT, feature probing on the vision tower (i.e., \(P_{(V, \text{end})}\)), and vision tower SFT trained via cross‑entropy.
As shown in Table~\ref{t3}, the vision tower SFT dramatically outperforms both MLLM SFT and feature probing across all datasets.
This stark discrepancy exposes the suboptimal state of the vision tower, which has been largely overlooked.
When optimized under the autoregressive text‑generation objective of MLLMs, the vision tower yields features far less discriminative than those obtained through the simple cross‑entropy loss of traditional supervised learning, ultimately capping the performance ceiling of medical image classification.

\begin{figure}[t]
\centerline{\includegraphics[width=\linewidth]{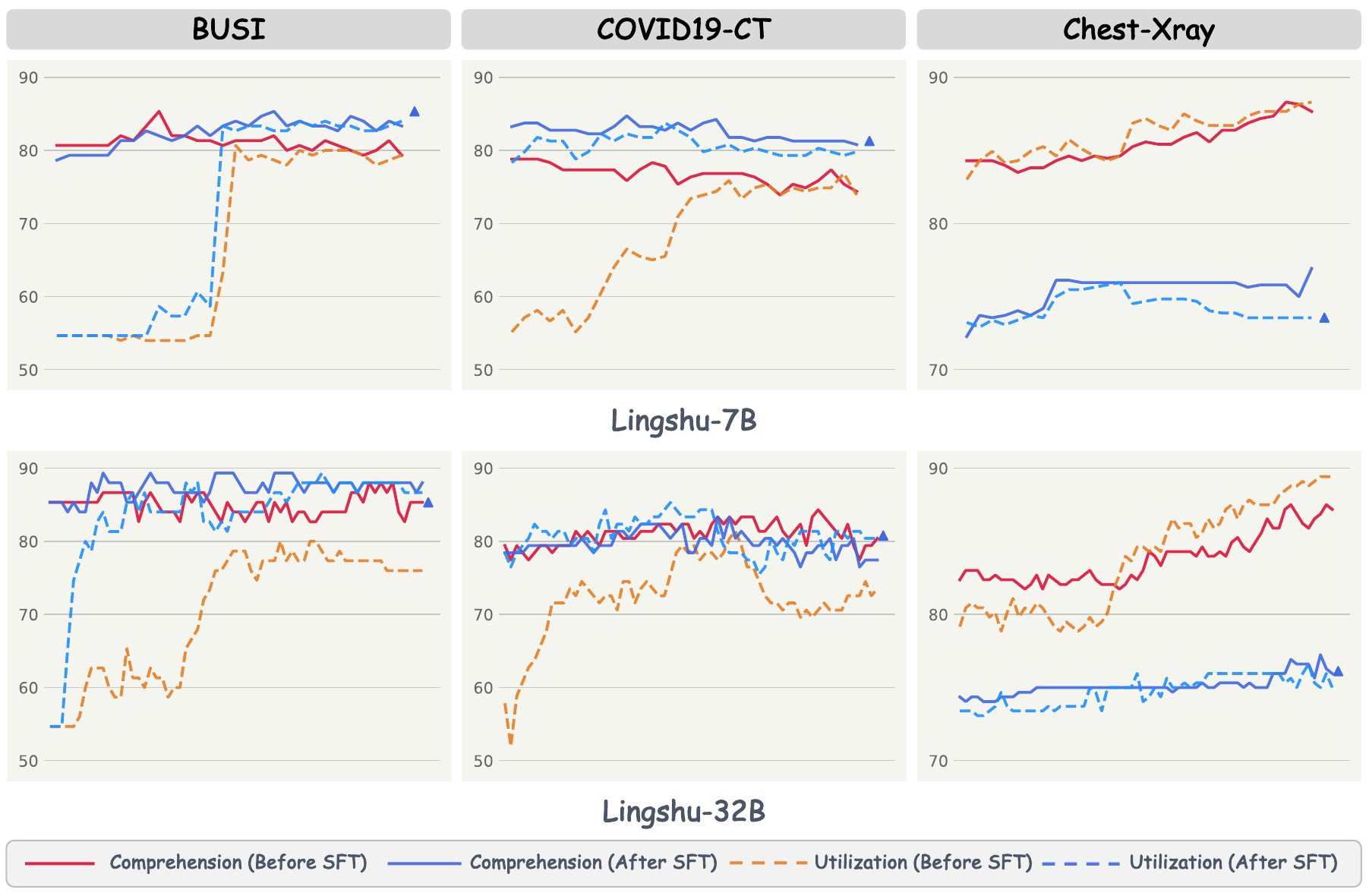}}
\caption{Comprehension and utilization dynamics in LLM reasoning. Layer-wise probing accuracy of Lingshu-7B and Lingshu-32B.}
\label{f5}
\end{figure}

\paragraph{The Comprehension-utilization Entanglement in LLM Reasoning.}
We dissect two internal capabilities of LLM reasoning: comprehension—the quality of visual representations encoded within the LLM layers, measured by probing averaged image token embeddings;
utilization—the ability to integrate all information at the decision point, measured by probing the final input token.
As illustrated in Figure~\ref{f5}, these two capabilities exhibit a tightly coupled hierarchical dependency.
MLLMs are trained via autoregressive next-token prediction, an objective that inherently optimizes the final token's representational utility rather than the quality of intermediate visual representations.
This training bias gives rise to a distinct temporal asynchrony: 
utilization curves typically originate below comprehension curves in early layers, but gradually catch up in mid-to-deep layers.
However, comprehension remains a prerequisite for utilization.
The ceiling of utilization (weighted aggregation) is fundamentally constrained by the ceiling of comprehension (averaged aggregation).
Consequently, when comprehension is severely limited—as observed on Chest-Xray—utilization struggles to develop even with fine-tuning.

\begin{table}[t]
\footnotesize
\tabcolsep= 0.15cm
\caption{Comparison of frozen vision tower (gray) vs. LoRA-finetuned vision tower on HuatuoGPT-Vision-7B and MedGemma-4B.}
\label{t2}
\begin{tabular}{cccccccc}
\toprule
\multirow{2}{*}{Models} & \multicolumn{2}{c}{BUSI} & \multicolumn{2}{c}{COVID19-CT} & \multicolumn{2}{c}{Chest-Xray} & Average \\
 & Acc & F1 & Acc & F1 & Acc & F1 & Gap \\
\midrule
\multirow{2}{*}{HuatuoGPT-Vision-7B} & 84.00 & 83.19 & 78.82 & 78.82 & 78.53 & 85.31 &  \\
 & \cellcolor[HTML]{D9D9D9}82.67 & \cellcolor[HTML]{D9D9D9}80.92 & \cellcolor[HTML]{D9D9D9}78.33 & \cellcolor[HTML]{D9D9D9}78.22 & \cellcolor[HTML]{D9D9D9}80.29 & \cellcolor[HTML]{D9D9D9}86.32 & \cellcolor[HTML]{D9D9D9}-0.38\% \\
\multirow{2}{*}{MedGemma-4B} & 84.00 & 83.34 & 80.79 & 80.98 & 70.51 & 80.91 &  \\
 & \cellcolor[HTML]{D9D9D9}83.33 & \cellcolor[HTML]{D9D9D9}82.51 & \cellcolor[HTML]{D9D9D9}78.82 & \cellcolor[HTML]{D9D9D9}78.39 & \cellcolor[HTML]{D9D9D9}74.84 & \cellcolor[HTML]{D9D9D9}83.24 & \cellcolor[HTML]{D9D9D9}+0.27\% \\
\bottomrule
\end{tabular}
\end{table}

\begin{table}[t]
\footnotesize
\tabcolsep= 0.03cm
\caption{Accuracy comparison of MLLM SFT, feature probing $\bm{P_{{(V, \text{end})}}}$, and vision tower SFT on ShizhenGPT-7B-VL and MedGemma-1.5-4B.}
\label{t3}
\begin{tabular}{ccccccc}
\toprule
\multirow{2}{*}{Training Paradigm} & \multicolumn{3}{c}{ShizhenGPT-7B-VL} & \multicolumn{3}{c}{MedGemma-1.5-4B} \\
 & BUSI & COVID19-CT & Chest-Xray & BUSI & COVID19-CT & Chest-Xray \\
\midrule
MLLM SFT & 86.00 & 78.82 & 73.72 & 89.33 & 83.74 & 70.03 \\
Feature Probing & 84.67 & 79.80 & 74.52 & 86.00 & 81.28 & 69.87 \\
Vision Tower SFT & \textbf{89.33} & \textbf{81.28} & \textbf{88.30} & \textbf{93.33} & \textbf{89.10} & \textbf{85.22} \\
\bottomrule
\end{tabular}
\end{table}

\section{Conclusions}
This paper systematically dissects why medical MLLMs underperform in image classification—a foundational task where their promised advantages have consistently fallen short.
Through layer-wise, module-level probing across 14 models and three datasets, we pinpoint where and how classification signals are distorted along the pipeline.
Our findings distill the observed weaknesses into four failure modes: quality limitation in visual representation, fidelity loss in connector projection, comprehension deficit in LLM reasoning, and misalignment of semantic mapping.
To enable principled comparisons, we introduce quantitative health scores that capture the evolution of visual representation across modules.
Beyond diagnosis, our discussions reveal three critical barriers.
First, medical adaptation offers only marginal gains over general-domain counterparts.
Second, vision towers trained under autoregressive objectives yield suboptimal classification representations compared to those trained with cross‑entropy supervision.
Third, comprehension and utilization of visual information in the LLM are tightly entangled, with the former fundamentally limiting the latter. 
We hope that this work serves as both a caution and a roadmap.
The road from hype to clinically reliable MLLMs remains long, and foundational capabilities such as classification deserve deliberate engineering rather than being taken for granted.

\bibliographystyle{ACM-Reference-Format}
\bibliography{sample-base}

\clearpage
\twocolumn[%
  \begin{@twocolumnfalse} 
    \begin{center}
      \Huge\bfseries Supplementary Materials: \\
        Lost in the Hype: Revealing and Dissecting the Performance Degradation of Medical Multimodal Large Language Models in Image Classification
    \end{center}
  \end{@twocolumnfalse}
  \vspace{20pt}
]

\appendix

\renewcommand{\thefigure}{S\arabic{figure}}
\renewcommand{\thetable}{S\arabic{table}}
\setcounter{figure}{0}  
\setcounter{table}{0}

\section{Instruction Template}
We provide one illustrative sample for each diagnostic category in Figures~\ref{fig:busi}, \ref{fig:covid}, and \ref{fig:chest}, covering the BUSI, COVID19-CT, and Chest-Xray datasets, respectively. 

\begin{tcolorbox}[title = {BUSI}]

    \begin{minipage}{\textwidth}
    \textbf{Instruction}: <image> Carefully examine the image to determine the diagnostic result for breast cancer. Please answer 'Normal', 'Benign', or 'Malignant'.
    \end{minipage}
    \smallskip 
    \begin{minipage}{\textwidth}
        \centering
        \includegraphics[width=0.6\textwidth]{supplementary_figures/busi1.png}
    \end{minipage}
    \smallskip 
    \begin{minipage}{\textwidth}
    \centering  
    \textbf{Ground truth}: Normal.
    \end{minipage}
    \smallskip  
    \begin{minipage}{\textwidth}
        \centering
        \includegraphics[width=0.6\textwidth]{supplementary_figures/busi2.png}
    \end{minipage}
    \smallskip  
    \begin{minipage}{\textwidth}
    \centering  
    \textbf{Ground truth}: Benign.
    \end{minipage}
    \begin{minipage}{\textwidth}
        \centering
        \includegraphics[width=0.6\textwidth]{supplementary_figures/busi3.png}
    \end{minipage}
    \smallskip  
    \begin{minipage}{\textwidth}
    \centering  
    \textbf{Ground truth}: Malignant.
    \end{minipage}

\end{tcolorbox}
\captionof{figure}{The instruction template and class-wise samples of the BUSI dataset.}
\label{fig:busi}

\begin{tcolorbox}[title = {COVID19-CT}]
    \begin{minipage}{\textwidth}
    \textbf{Instruction}: <image> Carefully examine the image to determine if there is COVID present. Please answer 'Yes' or 'No'.
    \end{minipage}
    \smallskip 
    \begin{minipage}[t]{0.48\textwidth}
        \centering
        \includegraphics[width=\linewidth]{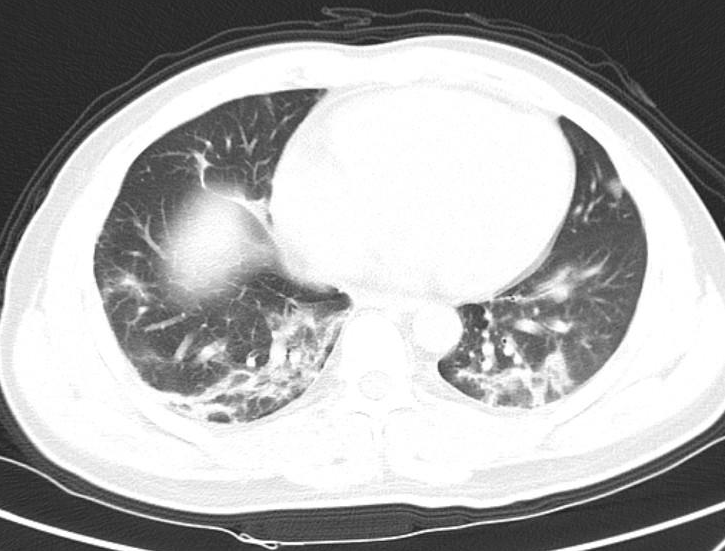}
        \smallskip
        \textbf{Ground truth}: Yes.
    \end{minipage}
    \hfill
    \begin{minipage}[t]{0.48\textwidth}
        \centering
        \includegraphics[width=\linewidth]{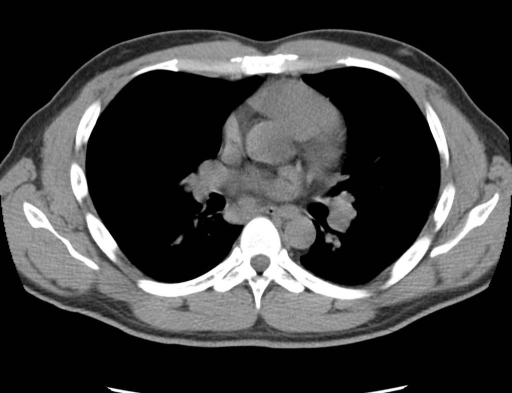}
        \smallskip
        \textbf{Ground truth}: No.
    \end{minipage}
\end{tcolorbox}
\captionof{figure}{The instruction template and class-wise samples of the COVID19-CT dataset.}
\label{fig:covid}

\begin{tcolorbox}[title = {Chest-Xray}]
    \begin{minipage}{\textwidth}
    \textbf{Instruction}: <image> Carefully examine the image to determine if there is pneumonia present. Please answer 'Yes' or 'No'.
    \end{minipage}
    \smallskip 
    
    \begin{minipage}[t]{0.48\textwidth}
        \centering
        \includegraphics[width=\linewidth]{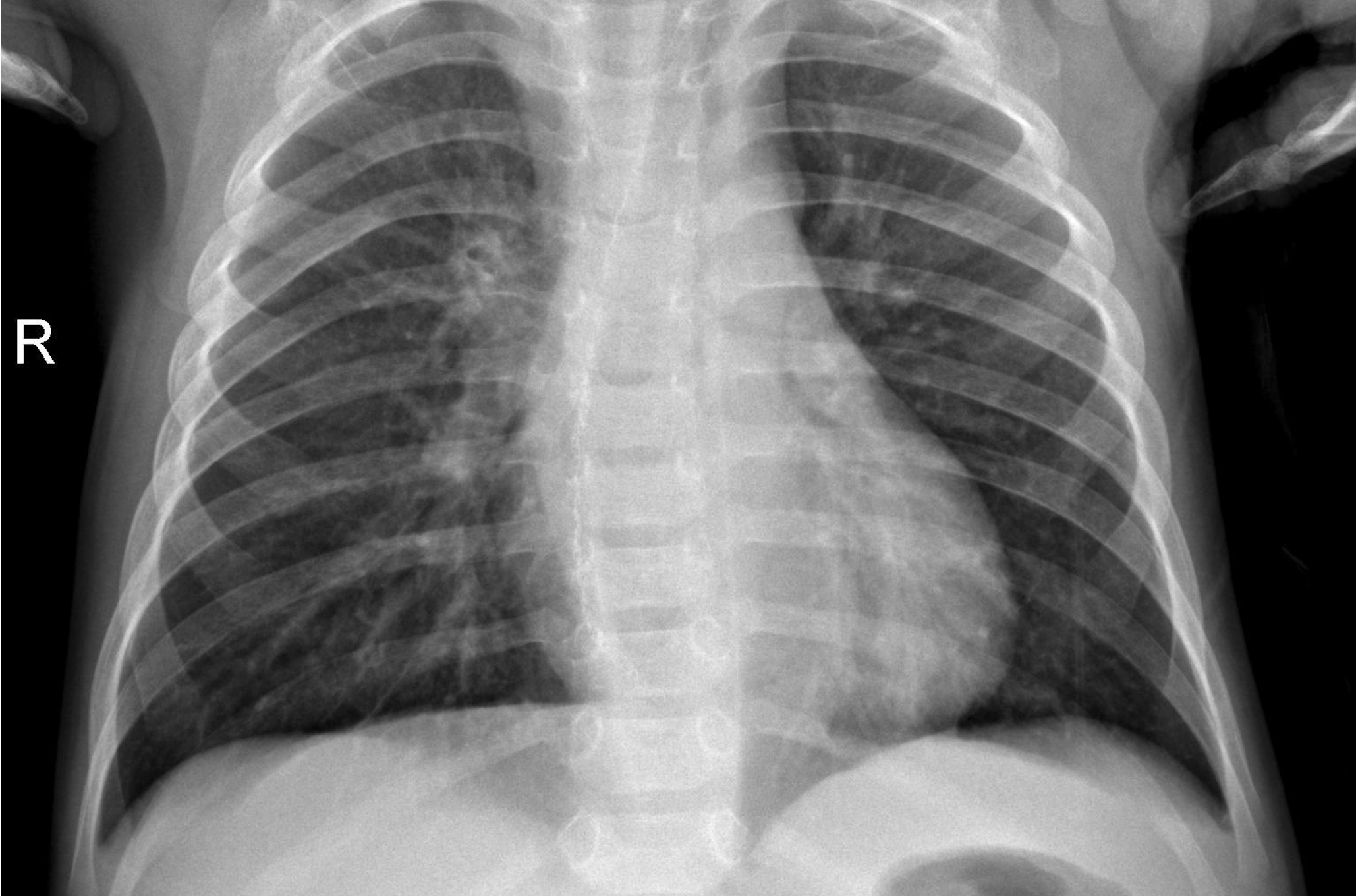}
        \smallskip
        \textbf{Ground truth}: Yes.
    \end{minipage}
    \hfill
    \begin{minipage}[t]{0.48\textwidth}
        \centering
        \includegraphics[width=\linewidth]{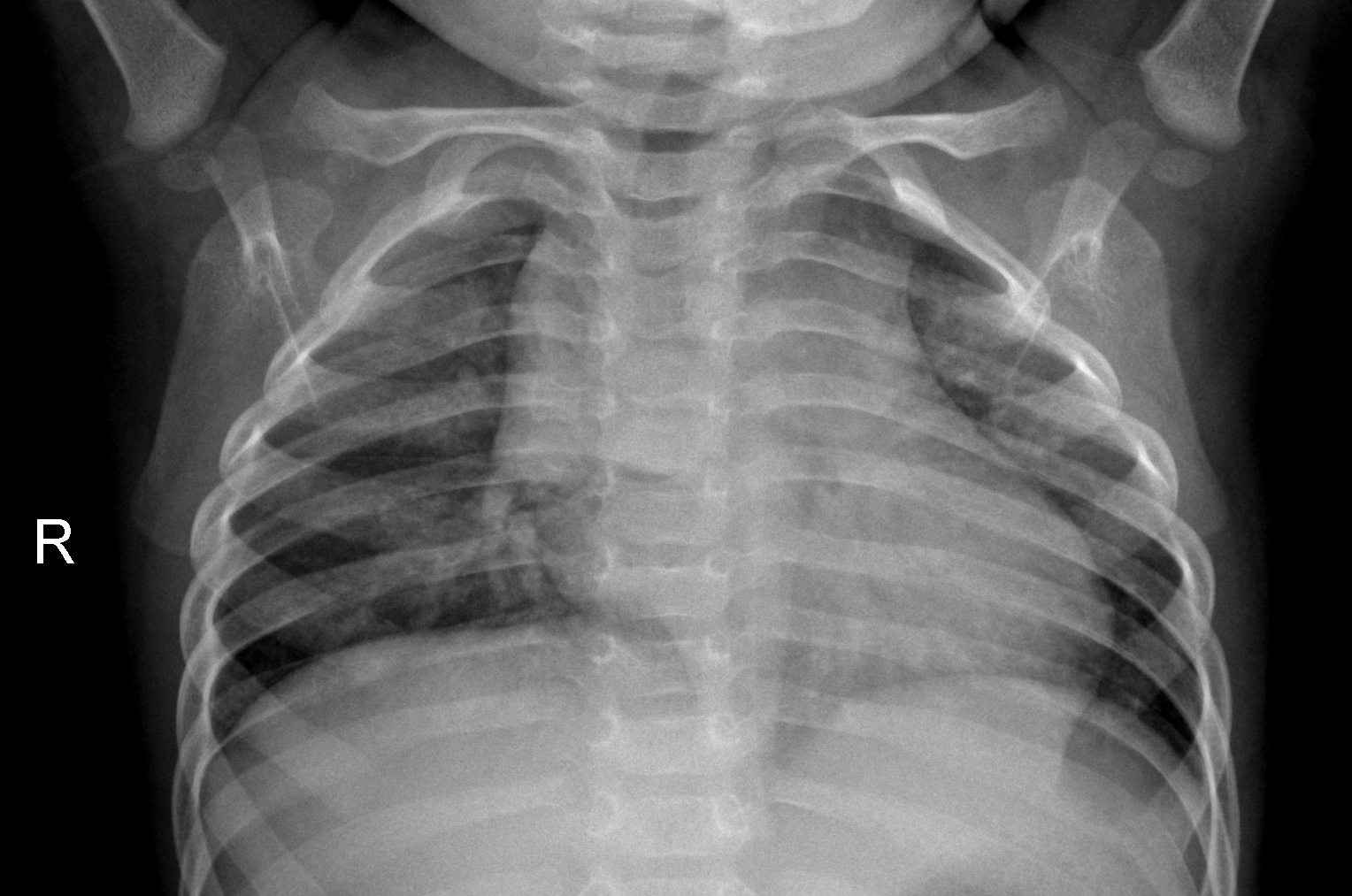}
        \smallskip
        \textbf{Ground truth}: No.
    \end{minipage}
\end{tcolorbox}
\captionof{figure}{The instruction template and class-wise samples of the Chest-Xray dataset.}
\label{fig:chest}

\section{Medical MLLM}

Tabel~\ref{a2t} provides detailed information for each medical MLLM, including the model name, parameter count, the base MLLM it initializes from, vision tower type, connector type, LLM type, model description, and training strategy.

\begin{table*}[p]
\tiny
\renewcommand{\arraystretch}{1.2}
\caption{Detailed information for each medical MLLM.}
\label{a2t}
\begin{tabular}{ccccccm{2cm}m{5cm}}
\toprule
\textbf{Medical MLLMs} & \textbf{Params} & \textbf{Base MLLM} & \textbf{Vision Tower} & \textbf{Connector} & \textbf{LLM} & \textbf{Description} & \textbf{Training Stages} \\
\midrule
MedVLM-R1 & 2.21B & Qwen2-VL & \makecell{ViT\\trained from scratch} & MLP & Qwen2-2B & A lightweight medical VLM trained with reinforcement learning using the GRPO algorithm, achieving strong generalization and high efficiency with only 600 training samples. & \textbf{Stage 1:} Uses GRPO on 600 MRI VQA samples (only final answers, no reasoning labels)  to optimize format and accuracy rewards, encouraging emergent reasoning without explicit supervision. No separate SFT or multi-stage alignment. \\
\midrule

\makecell{HuatuoGPT-Vision-7B\\ \\ \\HuatuoGPT-Vision-34B}
& \makecell{8.29B\\ \\ \\ 34.75B} & \makecell{Qwen2.5-VL\\ \\ \\ -} & \makecell{Redesigned ViT\\trained from scratch \\ \\CLIP ViT-L/14@336} & \makecell{MLP\\ \\ \\MLP} & \makecell{Qwen2.5-7B\\ \\ \\Yi-1.5-34B} & A medical MLLM trained on PubMedVision (1.3M medical VQA entries), featuring bilingual capabilities and enhanced medical knowledge from HuatuoGPT-II. & \textbf{Stage 1:} Visual-language alignment using 558K general data (LLaVA) and 647K medical data (PubMedVision Alignment VQA), training only the two-layer MLP projector while keeping the CLIP vision encoder and LLM frozen following the LLaVA-1.5 settings. \newline \textbf{Stage 2:} Instruction tuning using 658K general data (LLaVA) and 647K medical data (PubMedVision Instruction-Tuning VQA), updating both the MLP projector and the LLM to enhance instruction-following capabilities following the LLaVA-1.5 settings. \newline \textbf{Additional:} A further 348K bilingual data from a Chinese medical VQA dataset (translated from PubMedVision) and the medical text corpus from HuatuoGPT-II (exact data size not specified) are incorporated to enhance bilingual capability and medical knowledge. \\
\midrule

\makecell{Lingshu-7B\\ \\ \\ \\ \\Lingshu-32B} & \makecell{8.29B\\ \\ \\ \\ \\33.45B} & \makecell{Qwen2.5-VL\\ \\ \\ \\ \\Qwen2.5-VL} & \makecell{Redesigned ViT\\trained from scratch \\ \\ \\ Redesigned ViT\\trained from scratch} & \makecell{MLP\\ \\ \\ \\ \\MLP} & \makecell{Qwen2.5-7B\\ \\ \\ \\ \\Qwen2.5-32B} & A generalist medical foundation model that progressively infuses extensive medical knowledge and enhances reasoning capabilities through a meticulously curated multi-modal dataset and a multi-stage training paradigm. & \textbf{Stage 1:} Fine-tune the vision encoder and projector (LLM frozen) on \textasciitilde 927K medical image–caption pairs from PMC-OA and ROCO. \newline \textbf{Stage 2:} Perform end-to-end training of the vision encoder, projector, and LLM on \textasciitilde 4.1M image–text pairs, including medical caption datasets (ROCOv2, LLaVA-Med, Quilt-LLaVA, PubMedVision, MIMIC-CXR, FairVLMed, MedICaT, MedPix-2.0, synthesized long captions) and general caption datasets (LLaVA-1.5 Caption, PixMo). \newline \textbf{Stage 3:} Train the vision encoder, projector, and LLM on \textasciitilde 7.1M instruction samples consisting of medical multimodal instruction datasets (PathVQA, PMC-VQA, SLAKE, VQA-Med-2019, VQA-RAD, CheXpert Plus, IU-Xray, etc.), synthesized QA/OCR/CoT data, medical text instruction datasets (MedQA, MedQuAD, MedReason, MedThoughts-8K, ApolloCorpus, HealthCareMagic-100k, etc.), and general instruction data (LLaVA-1.5 instruct, ALLaVA, OpenHermes-2.5). \newline \textbf{Stage 4:} Apply GRPO-based RL with verifiable rewards on \textasciitilde 100K curated medical QA samples derived from datasets such as MIMIC-Ext-MIMIC-CXR-VQA, PMC-VQA, PathVQA, SLAKE, VQA-Med-2019, and VQA-RAD to improve medical reasoning ability. \\
\midrule

\makecell{ShizhenGPT-7B-VL\\ \\ \\ \\ \\ShizhenGPT-32B-VL} & \makecell{8.29B\\ \\ \\ \\ \\33.45B} & \makecell{Qwen2.5-VL\\ \\ \\ \\ \\Qwen2.5-VL} & \makecell{Redesigned ViT\\trained from scratch \\ \\ \\ \makecell{Redesigned ViT\\trained from scratch}} & \makecell{MLP\\ \\ \\ \\ \\MLP} & \makecell{Qwen2.5-7B\\ \\ \\ \\ \\Qwen2.5-32B} & The first multimodal LLM tailored for Traditional Chinese Medicine (TCM), capable of understanding images, sounds, smells, and pulse signals. Supports TCM's "Four Diagnostic Methods" (looking, listening/smelling, questioning, and pulse-taking). & \textbf{Stage 1:} Text-only pre-training on 11.9B tokens (6.3B TCM tokens from TCM Web Corpus and TCM Book Corpus + 5.6B general tokens from FineWeb-edu corpus) using full-parameter training of the LLM to inject TCM knowledge while preserving general capabilities. \newline \textbf{Stage 2:} Multimodal pre-training on 3.84B tokens (1.17B TCM image-text tokens, 0.69B general image-text tokens from ShareGPT-4V, 0.03B TCM audio-text tokens, 0.04B general audio tokens, and 1.75B resampled text tokens) jointly training the LLM, vision adapter, and signal adapter for multimodal alignment. \newline \textbf{Stage 3:} Instruction tuning on multimodal datasets (83,629 TCM text instructions, 65,033 vision instructions with 70,638 images, and signal instructions: 99,195 ECG, 57,957 speech, 4,101 pulse, 672 smell, 456 cough, 189 heartbeat) using full-parameter tuning to obtain the final ShizhenGPT. \\
\midrule

\makecell{Hulu-Med-4B \\ \\ Hulu-Med-7B \\ \\ Hulu-Med-14B \\ \\ Hulu-Med-32B} & \makecell{4.83B \\ \\ 8.04B \\ \\ 15.21B \\ \\ 33.21B}  & \makecell{- \\ \\ - \\ \\ - \\ \\ -} & \makecell{SigLIP with 2D RoPE \\ \\ SigLIP with 2D RoPE \\ \\ SigLIP with 2D RoPE \\ \\ SigLIP with 2D RoPE} & \makecell{MLP \\ \\ MLP \\ \\ MLP \\ \\ MLP} & \makecell{Qwen3-VL-4B \\ \\ Qwen2.5-7B \\ \\ Qwen3-14B \\ \\ Qwen2.5-32B} & A transparent, generalist medical VLM designed to unify language-only, 2D/3D vision-language, and video understanding within a single architecture. & \textbf{Stage 1:} Perform vision-language alignment by training the vision encoder and multimodal projector on 1.4M 2D medical image-caption pairs to align visual features with the LLM embedding space through short caption generation, while keeping the LLM frozen. \newline \textbf{Stage 2:} Perform medical multimodal pretraining by full-model training on 4.9M samples (public medical data and synthetic long captions) to inject medical knowledge and enhance visual understanding via long caption generation and open-ended QA. \newline \textbf{Stage 3:} Perform mixed-modality instruction tuning by continuing training all modules on 10.5M instruction samples (text and multimodal) to develop instruction-following and multi-task capabilities across text, 2D images, 3D volumes, and videos. \\
\midrule

\makecell{MedGemma-4B\\ \\ \\ \\ \\MedGemma-27B} & \makecell{4.30B\\ \\ \\ \\ \\27.43B} & \makecell{Gemma-3\\ \\ \\ \\ \\Gemma-3} & \makecell{MedSigLIP\\ \\ \\ \\ \\MedSigLIP} & \makecell{MLP\\ \\ \\ \\ \\MLP} & \makecell{Gemma-3-4B\\ \\ \\ \\ \\Gemma-3-27B} & A multimodal medical model built on Gemma-3 with the MedSigLIP vision encoder, designed to process medical images and text jointly for clinical reasoning tasks. & \textbf{Stage 1:} Perform vision encoder enhancement by training the vision tower on approximately 33M medical image–text pairs (including 32.5M internal histopathology patches, 231K MIMIC-CXR, 199K EyePACS, \textasciitilde 60K CT-US1, \textasciitilde 48K MRI-US1, 41K PMC medical images, and dermatology datasets) to obtain MedSigLIP. \newline \textbf{Stage 2:} Perform multimodal pretraining using Gemma-3 general multimodal data mixed with medical image–text data (\textasciitilde 33M, with medical data upweighted) while training the LLM and the multimodal connector. \newline \textbf{Stage 3:} Conduct post-training with medical QA datasets (\textasciitilde 400K samples) including MedMCQA, MedQA, PubMedQA, HealthSearchQA, LiveQA, AfriMed-QA, and synthetic QA, optimizing the LLM and connector. \newline \textbf{Additional:} The training methodology for the 27B variant was the same as for the 4B variant, with the addition of two training datasets: EHRQA, to improve the model's inherent EHR understanding, and ChestImaGenome, to enable anatomy localization on chest X-ray images. \\
\midrule

MedGemma-1.5-4B & 4.30B & Gemma-3 & MedSigLIP & MLP & Gemma-3-4B & MedGemma variant with expanded support for high-dimensional medical imaging, WSI, longitudinal medical imaging, anatomical localization, medical document understanding, and EHR understanding, plus improved accuracy on medical text reasoning and modest gains on standard 2D image interpretation over MedGemma 1 4B. & Relative to MedGemma-4B, it incorporates new training datasets, including CT-RATE, CT Dataset 1, MRI Dataset 1, WSI-Path, pathology datasets, MS-CXR-T, Mendeley Clinical Laboratory Test Reports, EHR datasets, and EHRNoteQA. \\
\bottomrule

\end{tabular}
\end{table*}

\section{Additional Probing Metrics}
\begin{figure*}[p]
\centering
\includegraphics[width=\linewidth]{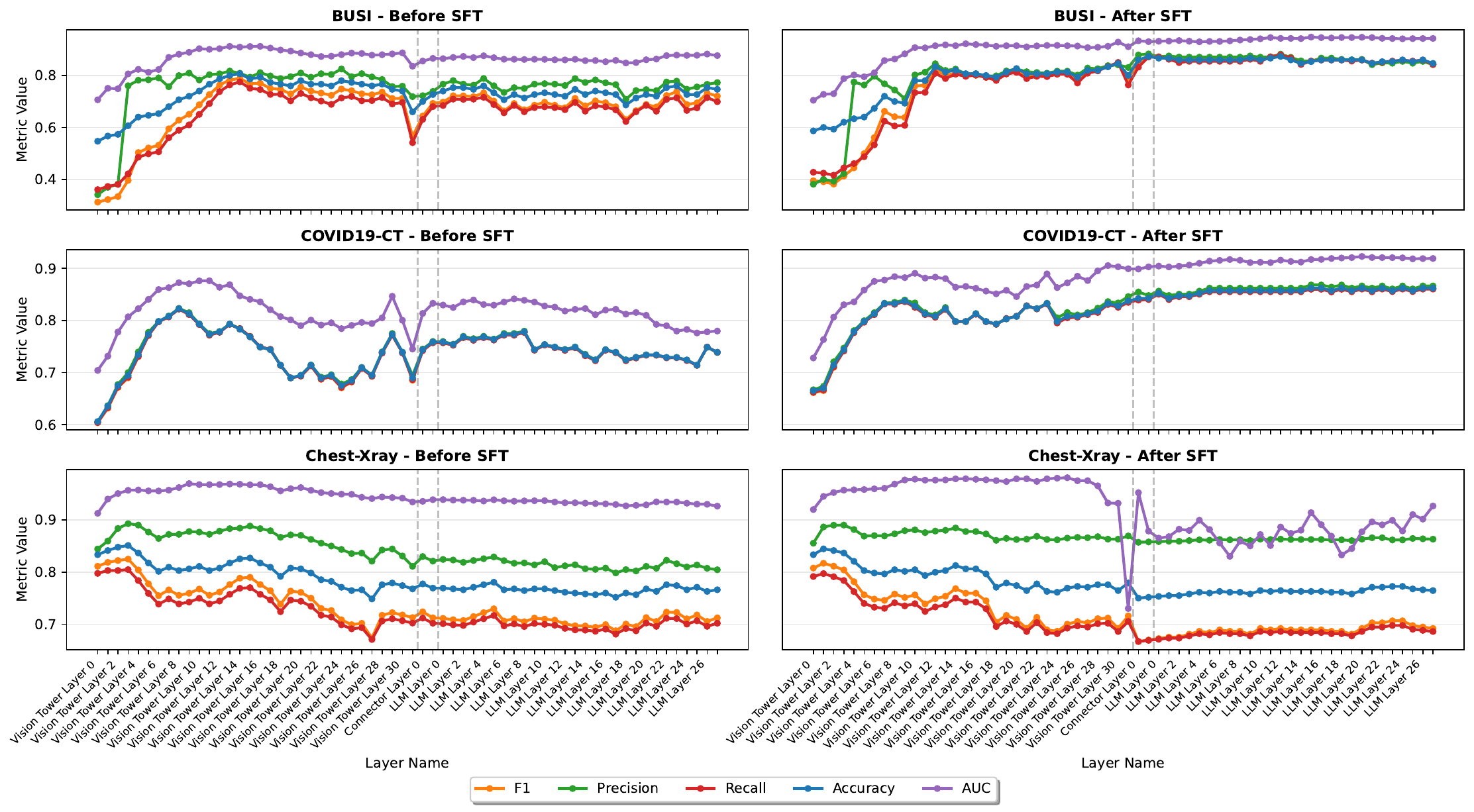}
\caption{Additional feature probing metrics of MedVLM-R1.}
\label{a3f1}
\end{figure*}

\begin{figure*}[p]
\centering
\includegraphics[width=\linewidth]{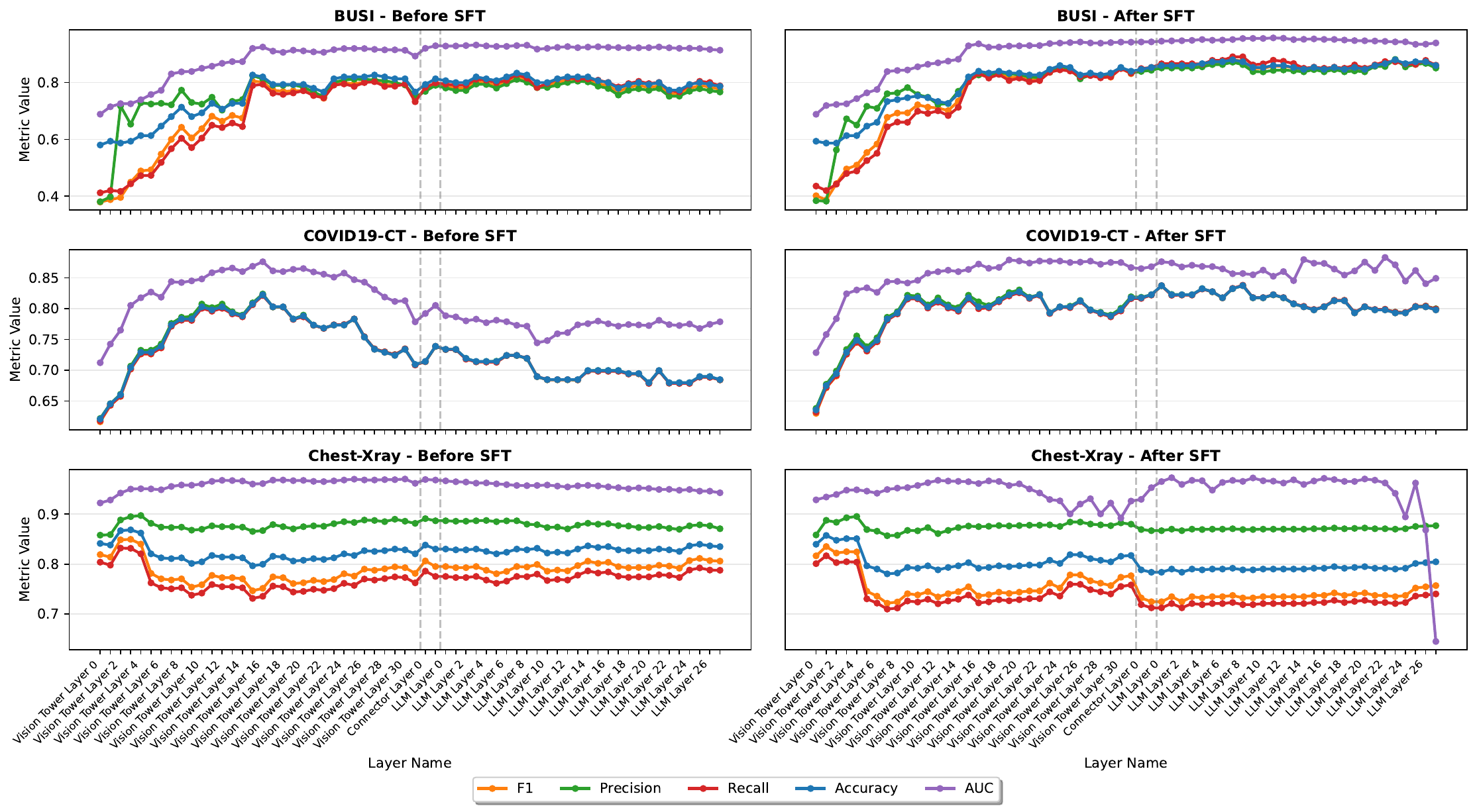}
\caption{Additional probing metrics of HuatuoGPT-Vision-7B.}
\label{a3f2}
\end{figure*}

\begin{figure*}[p]
\centering
\includegraphics[width=\linewidth]{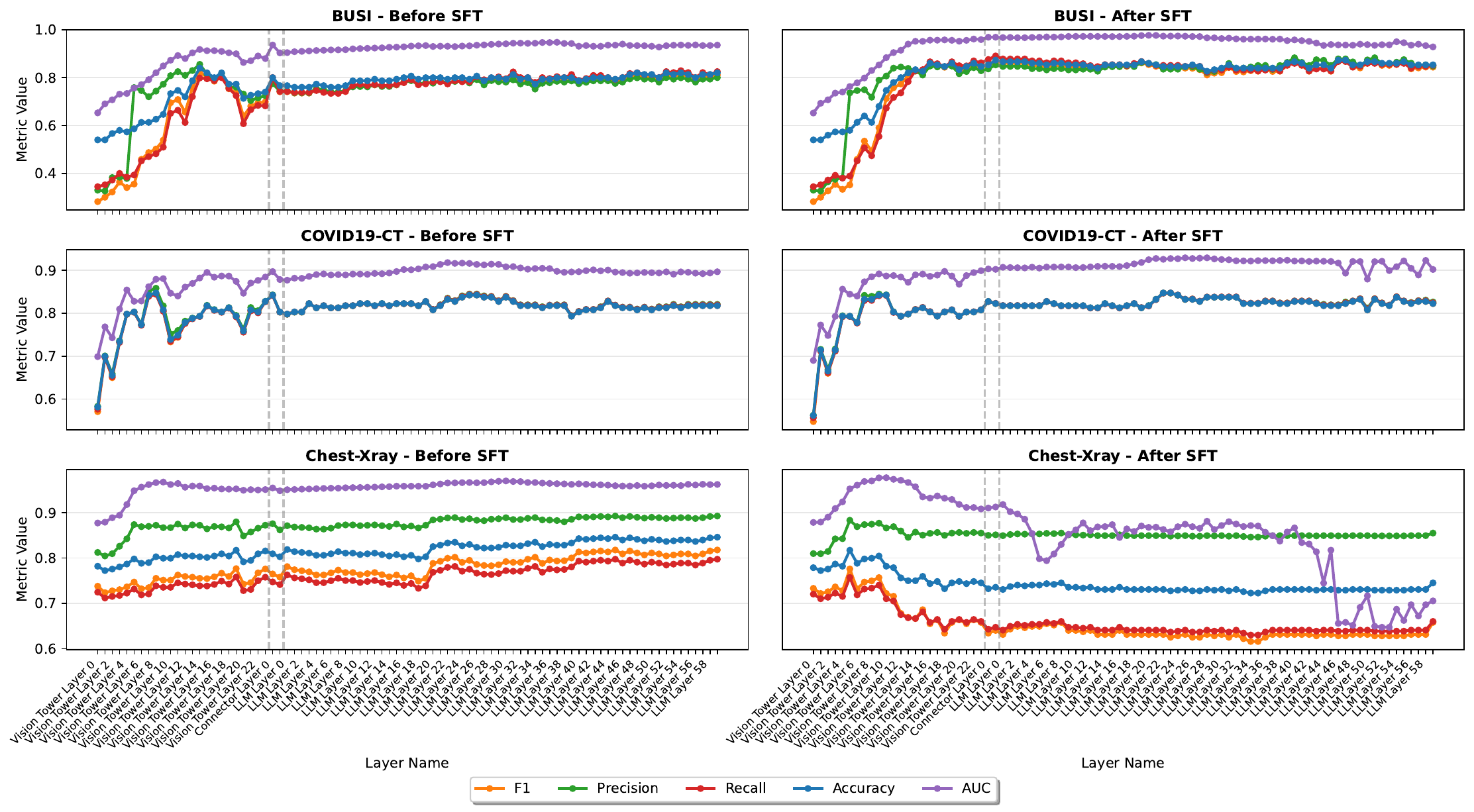}
\caption{Additional probing metrics of HuatuoGPT-Vision-34B.}
\label{a3f3}
\end{figure*}

\begin{figure*}[p]
\centering
\includegraphics[width=\linewidth]{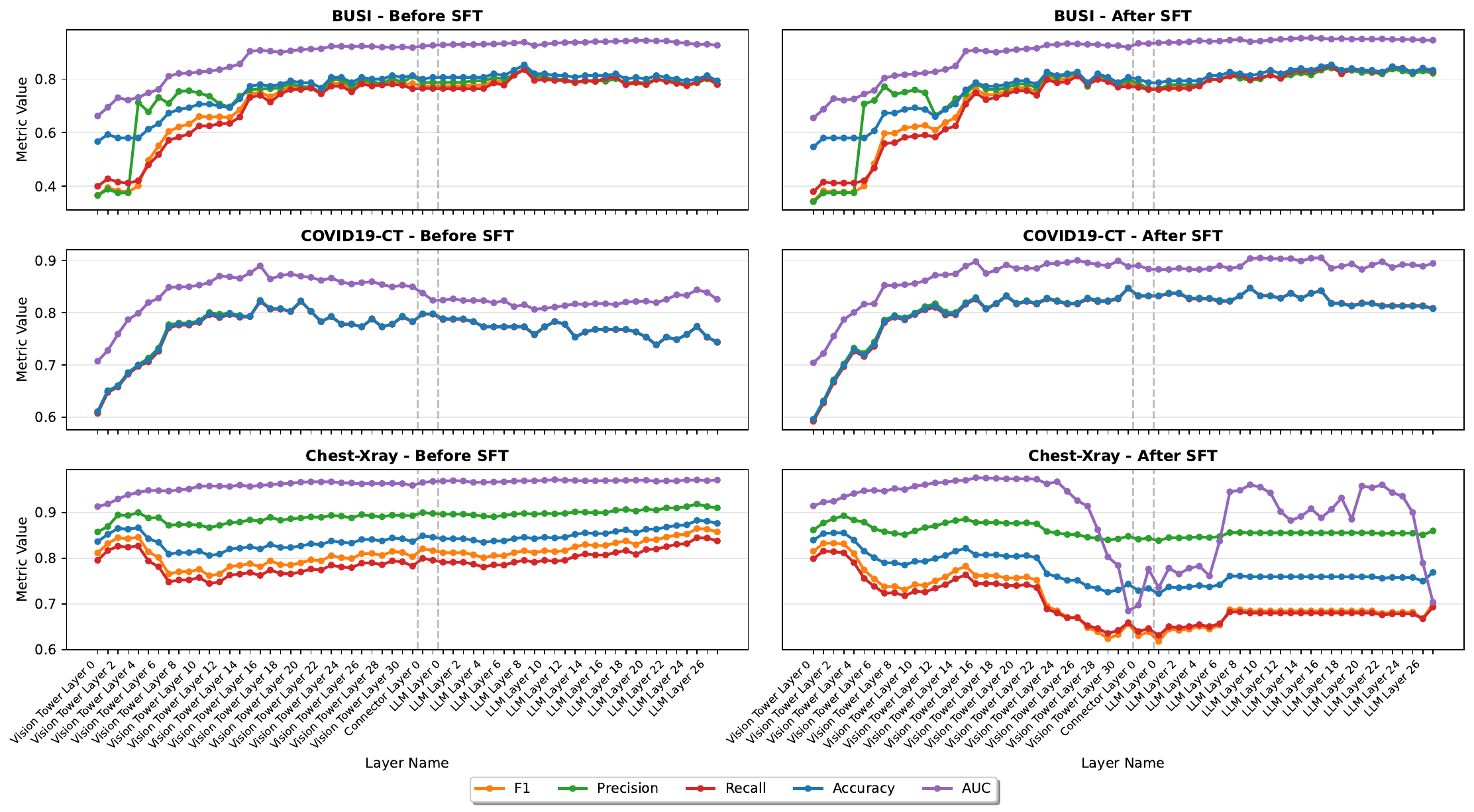}
\caption{Additional probing metrics of Lingshu-7B.}
\label{a3f4}
\end{figure*}

\begin{figure*}[p]
\centering
\includegraphics[width=\linewidth]{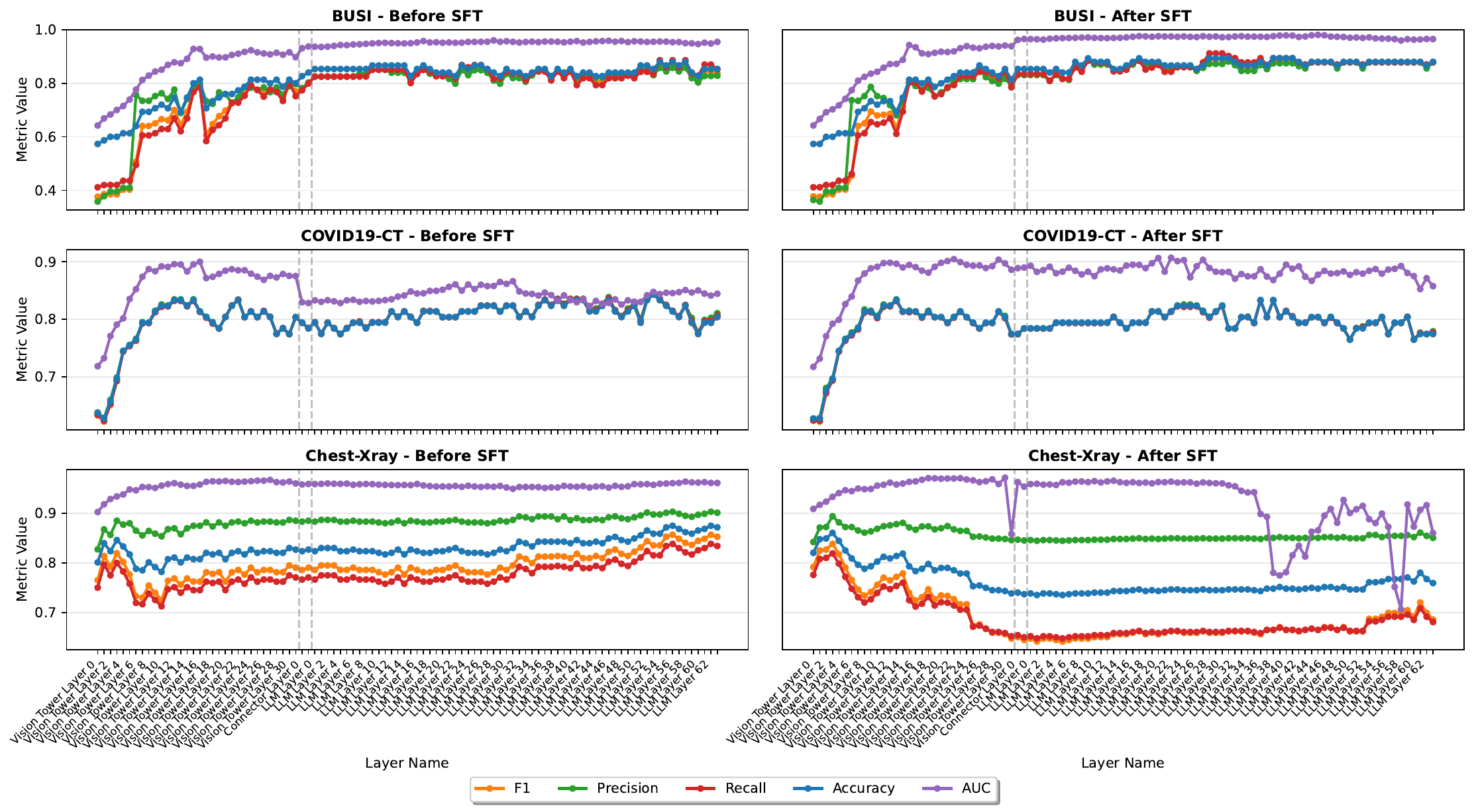}
\caption{Additional probing metrics of Lingshu-32B.}
\label{a3f5}
\end{figure*}

\begin{figure*}[p]
\centering
\includegraphics[width=\linewidth]{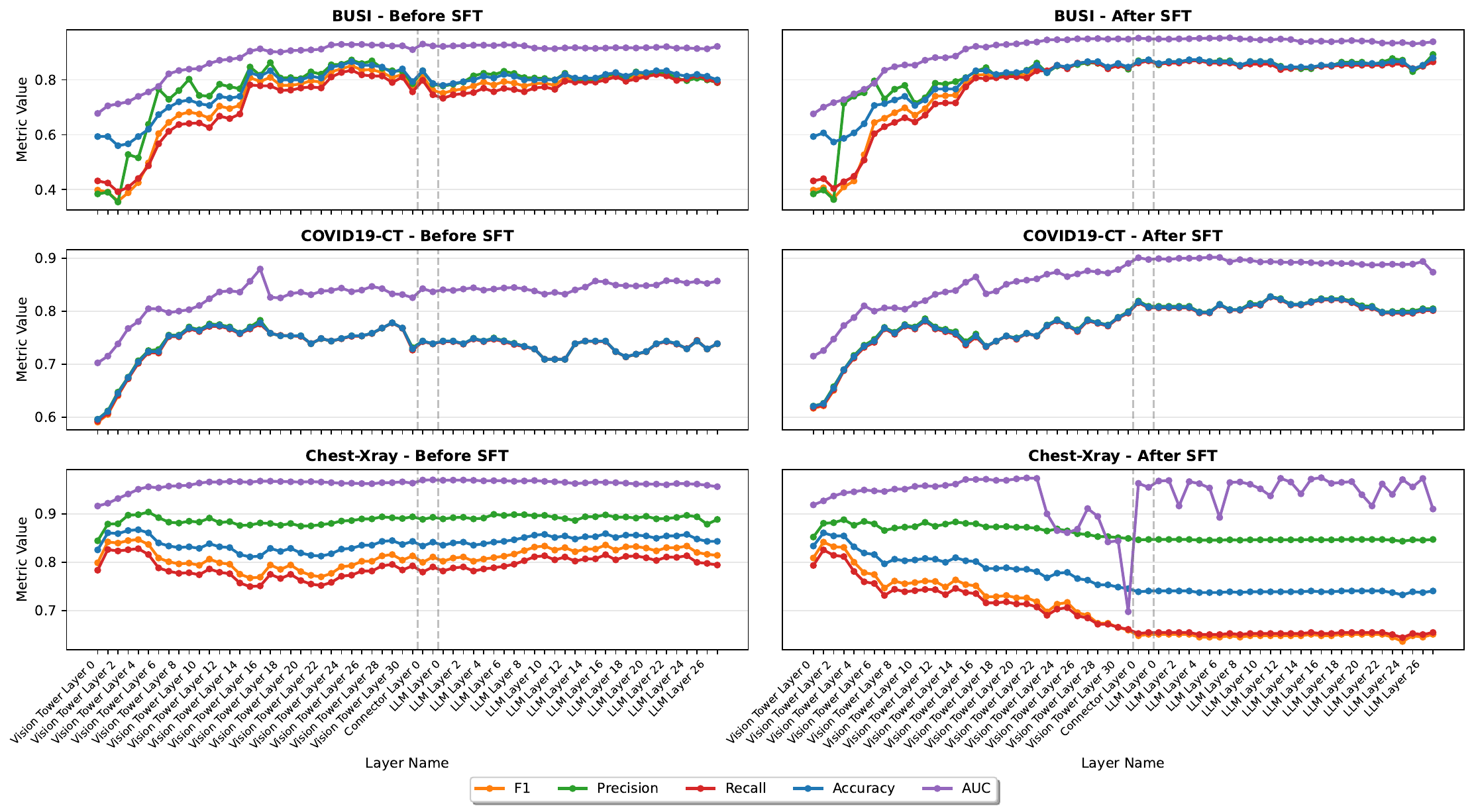}
\caption{Additional probing metrics of ShizhenGPT-7B-VL.}
\label{a3f6}
\end{figure*}

\begin{figure*}[p]
\centering
\includegraphics[width=\linewidth]{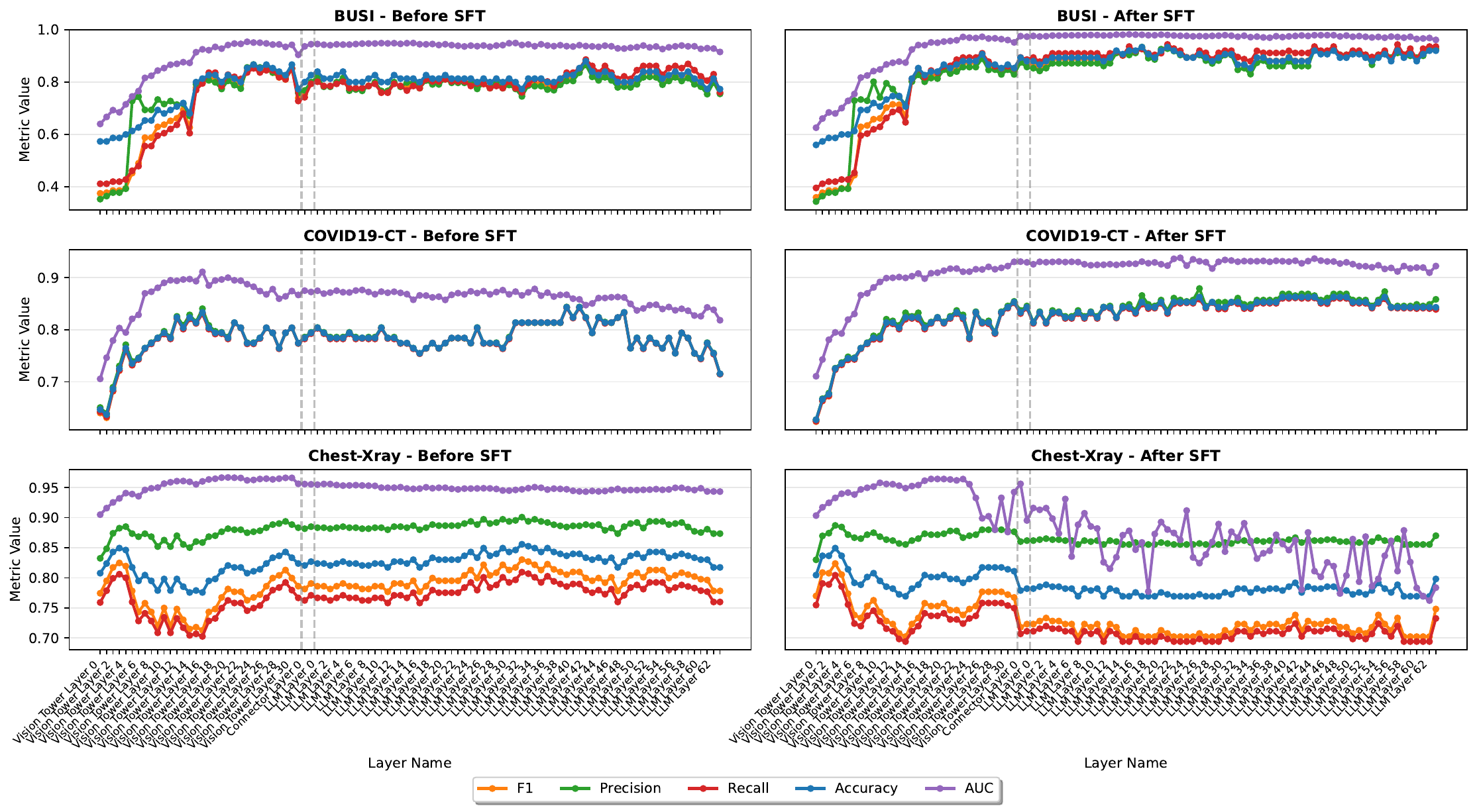}
\caption{Additional probing metrics of ShizhenGPT-32B-VL.}
\label{a3f7}
\end{figure*}

\begin{figure*}[p]
\centering
\includegraphics[width=\linewidth]{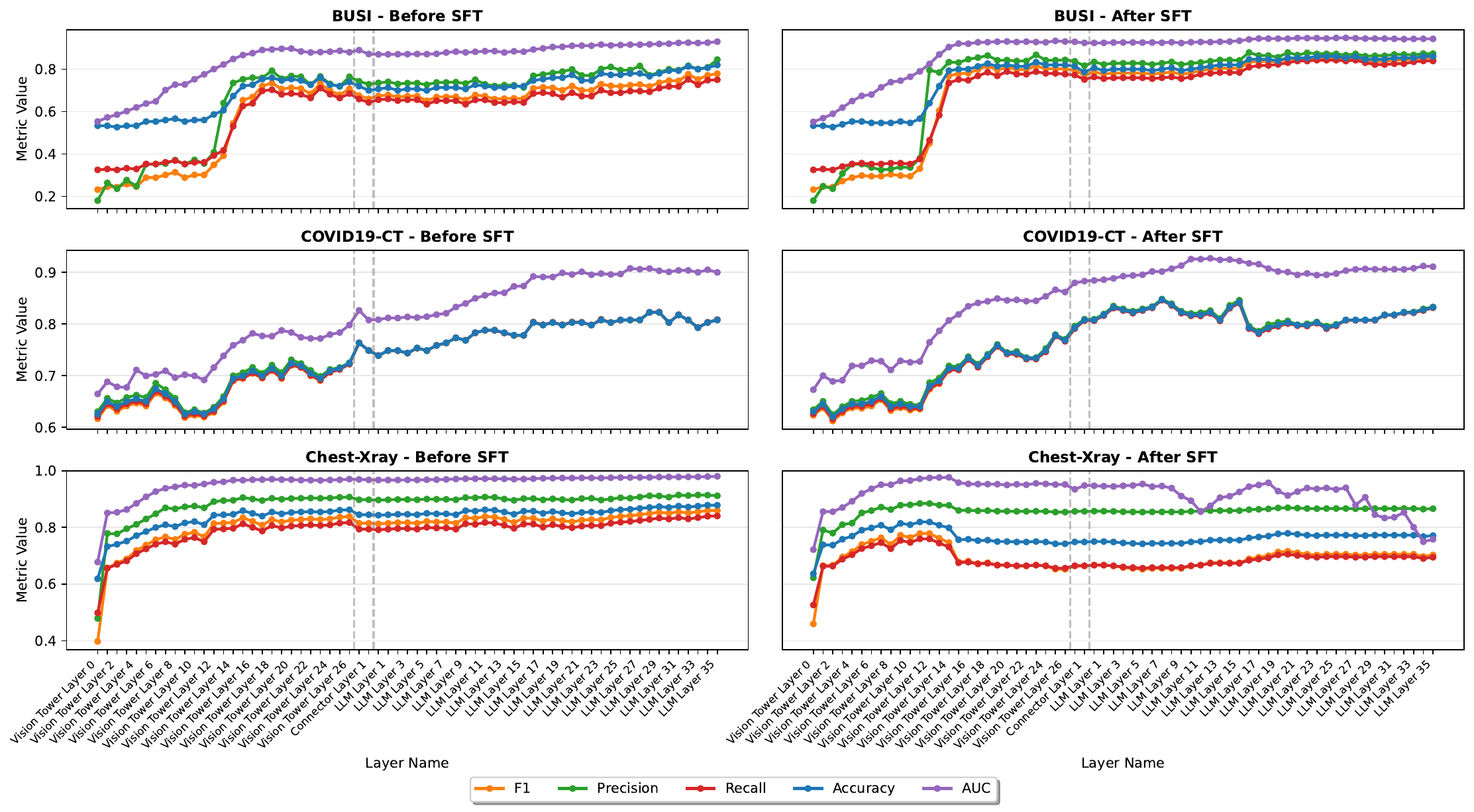}
\caption{Additional probing metrics of Hulu-Med-4B.}
\label{a3f8}
\end{figure*}

\begin{figure*}[p]
\centering
\includegraphics[width=\linewidth]{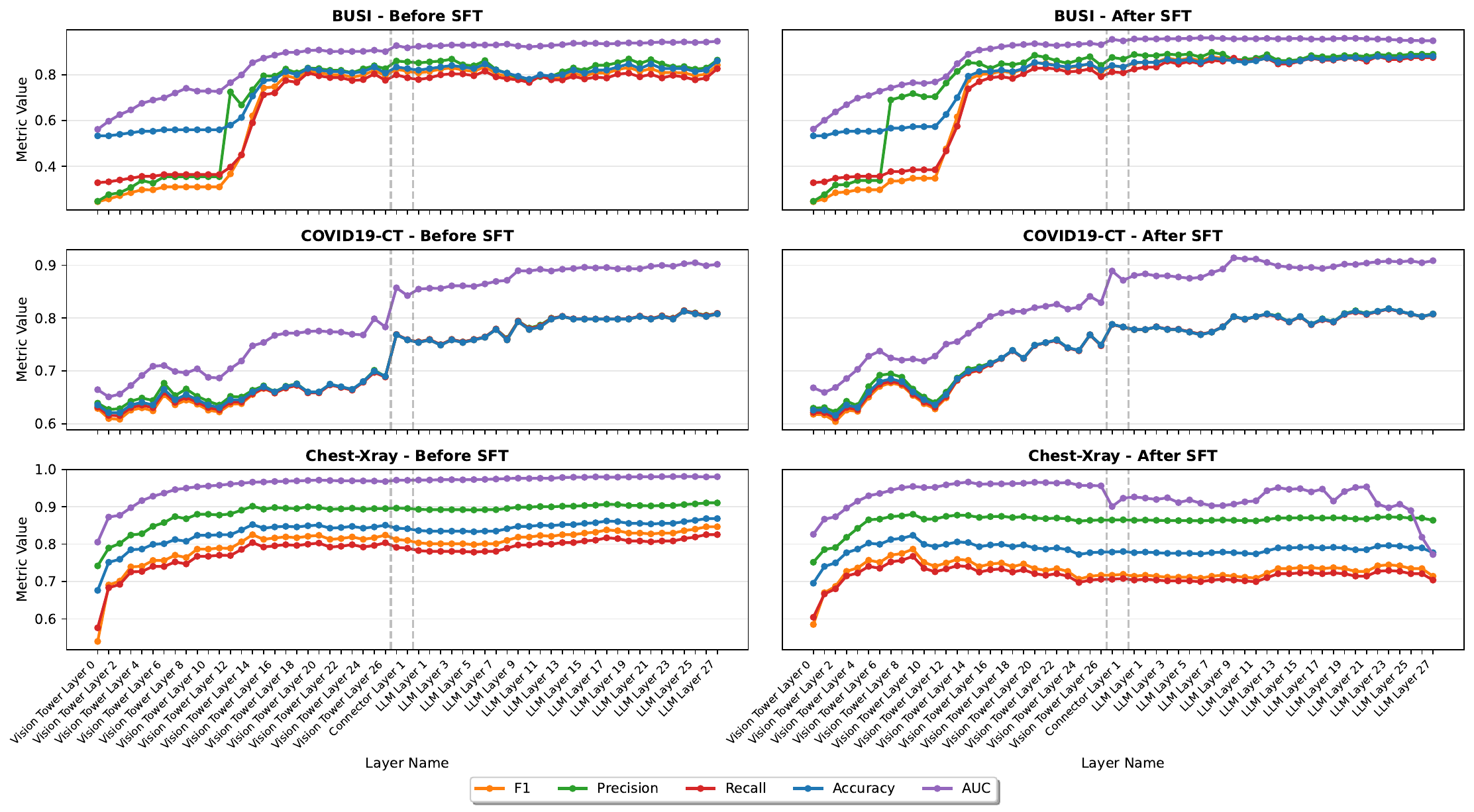}
\caption{Additional probing metrics of Hulu-Med-7B.}
\label{a3f9}
\end{figure*}

\begin{figure*}[p]
\centering
\includegraphics[width=\linewidth]{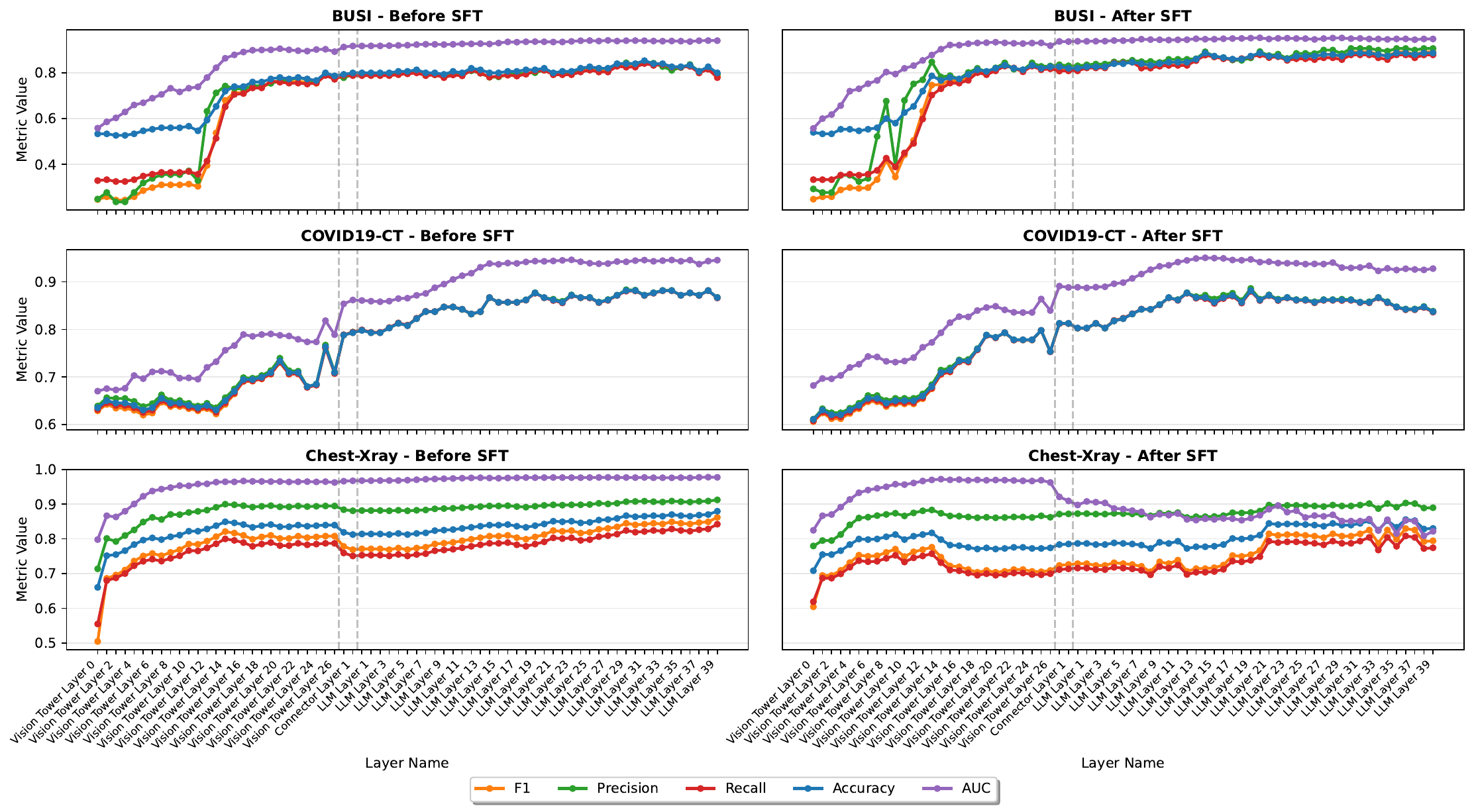}
\caption{Additional probing metrics of Hulu-Med-14B.}
\label{a3f10}
\end{figure*}

\begin{figure*}[p]
\centering
\includegraphics[width=\linewidth]{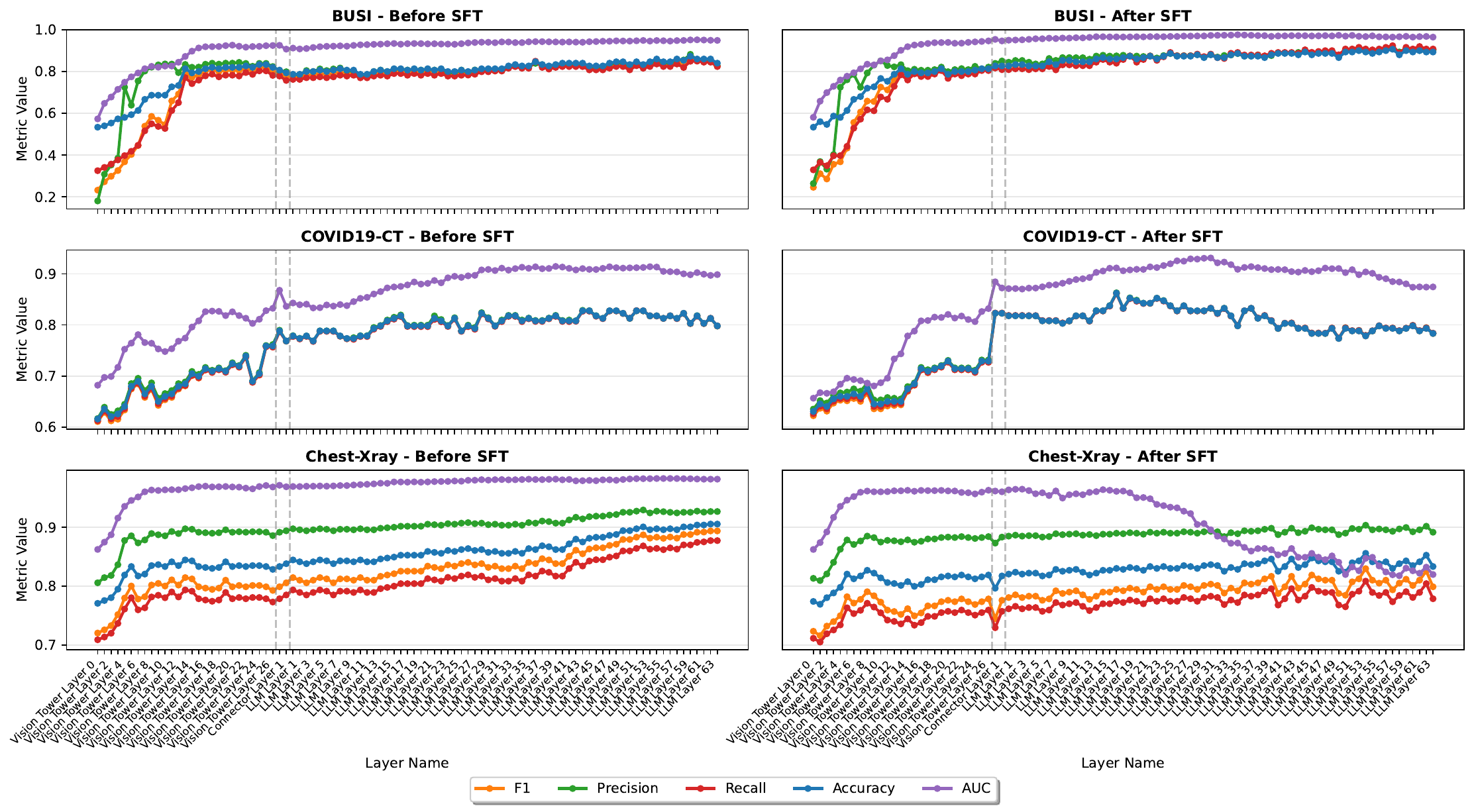}
\caption{Additional probing metrics of Hulu-Med-32B.}
\label{a3f11}
\end{figure*}

\begin{figure*}[p]
\centering
\includegraphics[width=\linewidth]{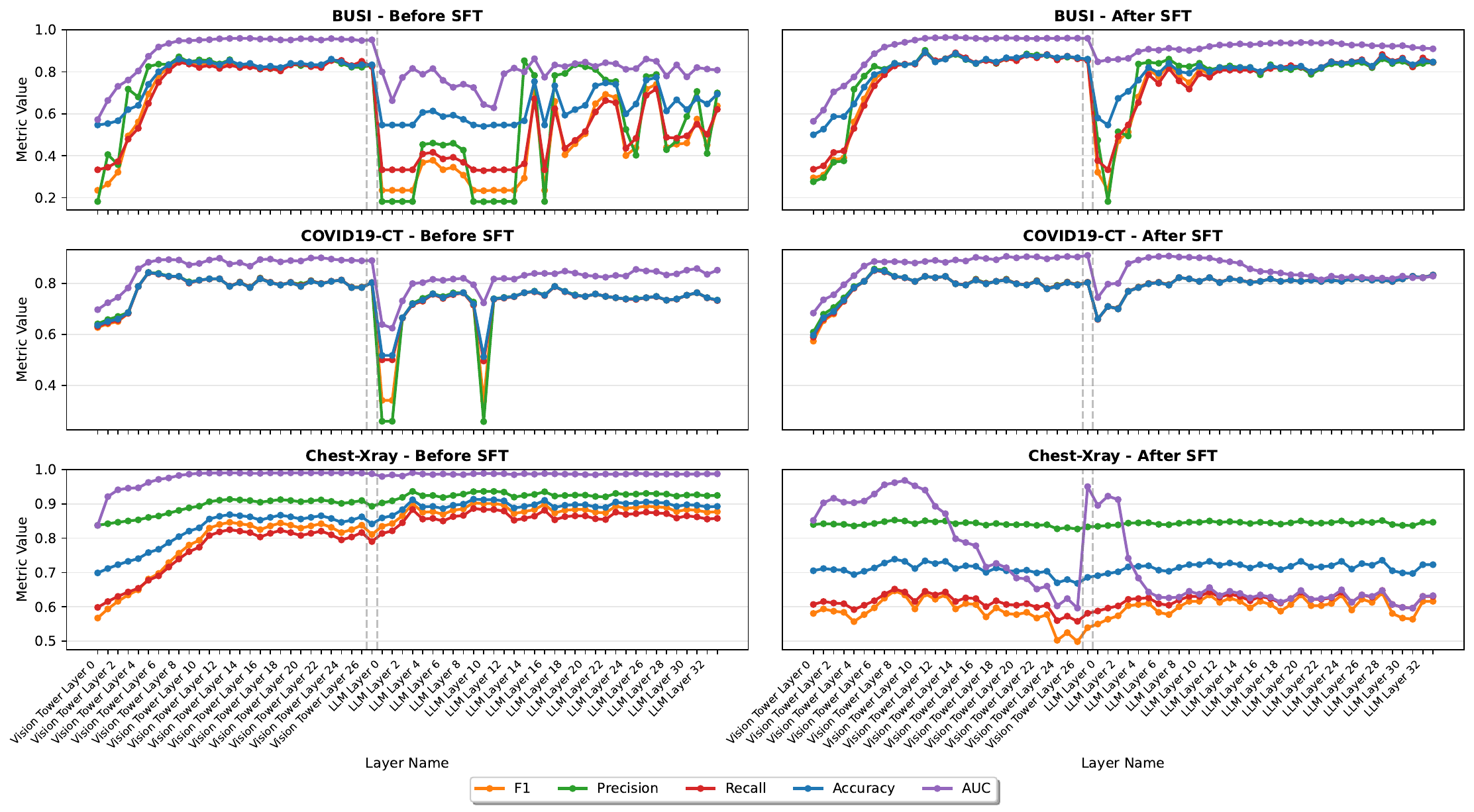}
\caption{Additional probing metrics of MedGemma-4B.}
\label{a3f12}
\end{figure*}

\begin{figure*}[p]
\centering
\includegraphics[width=\linewidth]{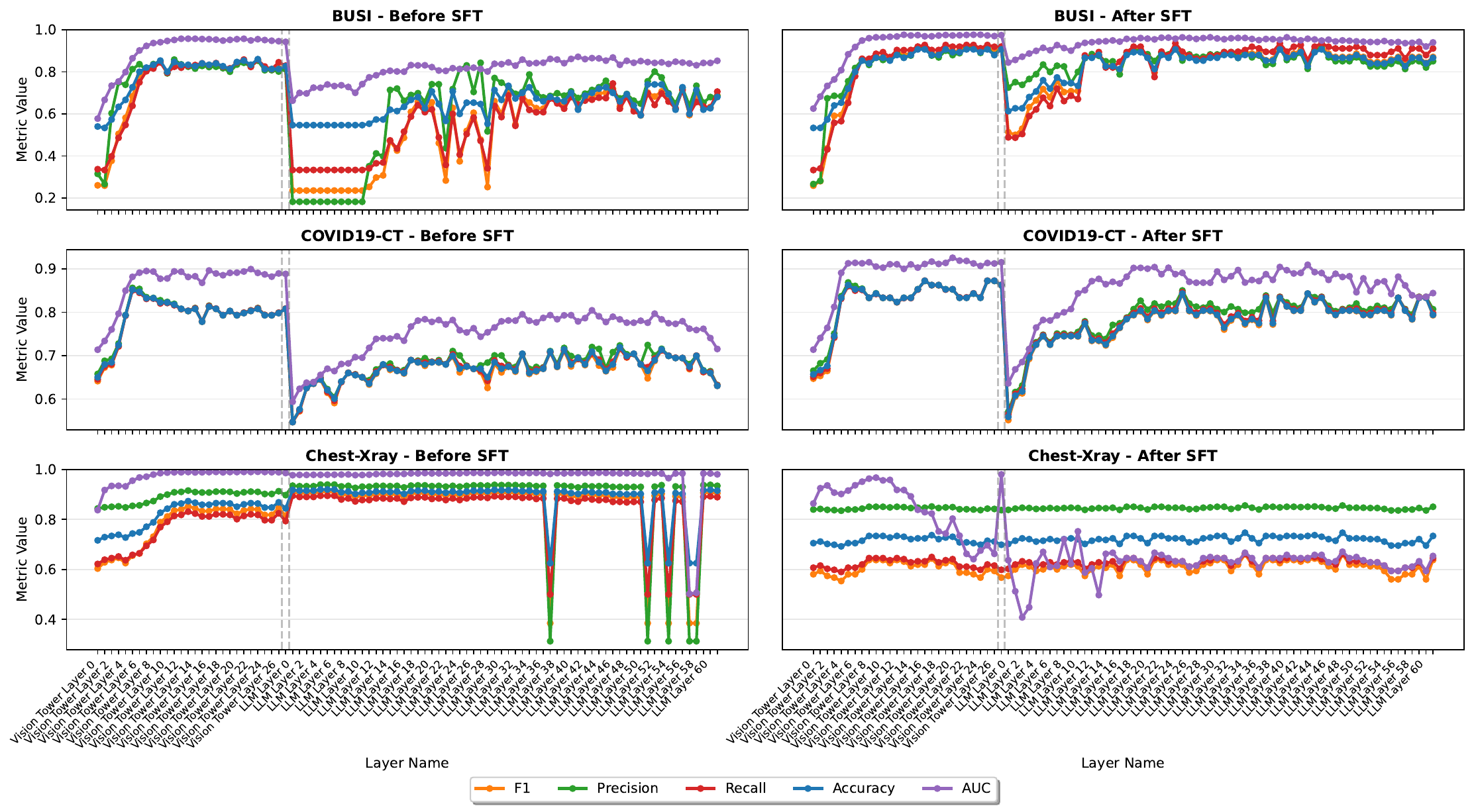}
\caption{Additional probing metrics of MedGemma-27B.}
\label{a3f13}
\end{figure*}

\begin{figure*}[p]
\centering
\includegraphics[width=\linewidth]{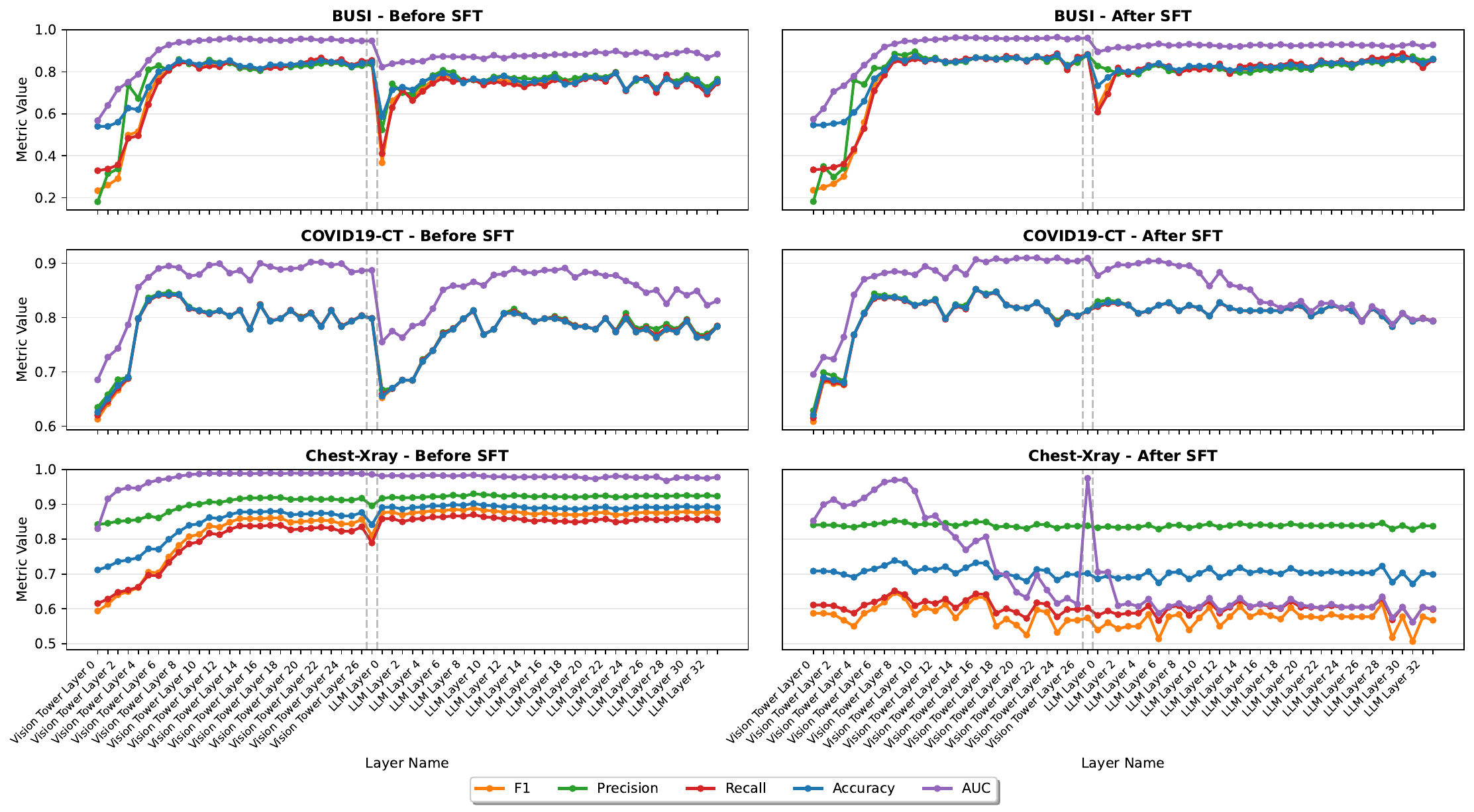}
\caption{Additional probing metrics of MedGemma-1.5-4B.}
\label{a3f14}
\end{figure*}

We provide comprehensive layer-wise feature probing results for all MLLMs in Figures~\ref{a3f1}--\ref{a3f14}, comparing model performance before and after SFT. Reported metrics include F1 score, precision, recall, accuracy, and AUC.

\end{document}